\documentclass{article}%
\usepackage{iclr2027_conference,times}
\iclrfinalcopy
\usepackage{listings}
\usepackage{float}
\usepackage{subcaption}

\usepackage[utf8]{inputenc}%
\usepackage[T1]{fontenc}%
\usepackage{hyperref}%
\usepackage{url}%
\usepackage{booktabs}%
\usepackage{amsfonts}%
\usepackage{nicefrac}%
\usepackage{silence}%
\usepackage{microtype}%
\usepackage{xcolor}%
\usepackage[flushmargin,hang]{footmisc}
\usepackage{hyperref}%
\hypersetup{
  colorlinks=true,
  linkcolor=blue!35!black,%
  citecolor=blue!35!black,%
  urlcolor=blue!35!black%
}

\usepackage{inconsolata}

\usepackage[greek.ancient,english]{babel}

\usepackage{hyperref}%
\usepackage{url}%
\usepackage{booktabs}%
\usepackage{amsfonts}%
\usepackage{nicefrac}%
\usepackage{microtype}%

\usepackage{latexsym}
\usepackage{amsfonts}
\usepackage{amsmath}
\usepackage{amssymb}
\usepackage{amsthm}

\usepackage{thmtools}
\usepackage{thm-restate}

\usepackage{relsize}
\usepackage{mathtools}
\usepackage{bm}
\usepackage{bbm}
\usepackage{xfrac}
\usepackage{stmaryrd}
\SetSymbolFont{stmry}{bold}{U}{stmry}{m}{n}
\usepackage[makeroom]{cancel}

\usepackage{xspace}
\usepackage{adjustbox}
\usepackage{relsize}

\usepackage{url}

\usepackage{array}

\usepackage[inline,shortlabels]{enumitem}
\setitemize{noitemsep,topsep=0pt,parsep=0pt,partopsep=0pt,leftmargin=*}
\setenumerate{noitemsep,topsep=0pt,parsep=0pt,partopsep=0pt,leftmargin=*}

\usepackage{caption}

\usepackage[normalem]{ulem}

\newtheorem{lemma}{Lemma}[section]

\theoremstyle{definition}
\newtheorem{example}{Example}[section]

\AtBeginEnvironment{example}{\pushQED{\hfill //}}
\AtEndEnvironment{example}{\popQED\endexample}

\usepackage{cleveref}
\crefname{section}{\S}{\S\S}
\crefname{table}{Tab.}{Tabs.}
\crefname{figure}{Fig.}{Figs.}
\crefname{algorithm}{Alg.}{Algs.}
\crefname{equation}{Eq.}{Eqs.}
\crefname{example}{Ex.}{Exs.}
\crefname{fact}{Fact}{Facts}
\crefname{appendix}{App.}{Appendices}
\crefname{theorem}{Thm.}{Thms.}
\crefname{reTheorem}{Thm.}{Thms.}
\crefname{aquestion}{Question}{Questions}
\crefname{assumption}{Assumption}{Assumptions}
\crefname{lemma}{Lem.}{Lemmas}
\crefname{reLemma}{Lem.}{Lemmas}
\crefname{proposition}{Prop.}{Props.}
\crefname{chapter}{Chapter}{Chapters}
\crefname{line}{line}{lines}
\crefname{principle}{Principle}{Principles}
\crefname{definition}{Def.}{Defs.}
\crefname{corollary}{Cor.}{Cors.}
\crefname{Exercise}{Exercise}{Exercises}
\crefformat{section}{\S#2#1#3}
\crefformat{footnote}{#2Footnote~#1#3}

\usepackage{colortbl}

\definecolor{ETHPetrol}{RGB}{0,120,148}%

\colorlet{MacroColor}{ETHPetrol}
\colorlet{MACROCOLOR}{MacroColor}

\usepackage{amssymb}

\newcommand{\inlineitem}[1][]{%
\ifnum\enit@type=\tw@
    {\descriptionlabel{#1}}
  \hspace{\labelsep}%
\else
  \ifnum\enit@type=\z@
       \refstepcounter{\@listctr}\fi
    \quad\@itemlabel\hspace{\labelsep}%
\fi}

\usepackage{algorithm}
\usepackage[noend]{algpseudocode}%
\algrenewcommand\algorithmicindent{1.0em}%

\newcommand{\rightcomment}[1]{{\color{gray} \(\triangleright\){\footnotesize\textit{#1}}}}
\algrenewcommand{\algorithmiccomment}[1]{\hfill \rightcomment{#1}}%
\algnewcommand{\LinesComment}[1]{\State {\color{black!50!green}\rightcomment{\parbox[t]{.95\linewidth-\leftmargin-\widthof{\(\triangleright\) }}{#1}}}}
\algnewcommand{\LineComment}[1]{\State {\color{black!50!green}\smaller \(\triangleright\) \parbox[t]{\linewidth-\leftmargin-\widthof{\(\triangleright\) }}{\it #1}\smallskip}}%
\algnewcommand{\InlineComment}[1]{\hfill {\color{black!50!green}\(\triangleright\) {\scriptsize \it #1}}}
\algrenewcommand\algorithmicindent{1.0em}%

\algrenewcommand\alglinenumber[1]{{\tiny\color{black!50}#1.}\hspace{-2pt}}
\newcommand{\algorithmicfunc}[1]{\textbf{def} {#1}:}
\algdef{SE}[FUNC]{Func}{EndFunc}[1]{\algorithmicfunc{#1}}{}
\makeatletter
\ifthenelse{\equal{\ALG@noend}{t}}%
{\algtext*{EndFunc}}
{}%
\makeatother

\usepackage{multicol}

\colorlet{MacroColor}{ETHPetrol}
\newcommand{\mymacro}[1]{{\color{MacroColor} #1}}

\newcommand{\defn}[1]{\textbf{#1}}
\newcommand{\defeq}{\mathrel{\stackrel{\textnormal{\tiny def}}{=}}}

\newlength\myheight
\newlength\mydepth
\settototalheight\myheight{Xygp}
\newcommand{\alphabet}{{\mymacro{\Sigma}}}
\newcommand{\str}{{\mymacro{\boldsymbol{x}}}}
\newcommand{\sym}{{\mymacro{x}}}

\newcommand{\func}[0]{\mymacro{\ensuremath{f}}\xspace}
\newcommand{\funcg}[0]{\mymacro{g}}

\newcommand{\p}{\ensuremath{\mymacro{p}}\xspace}
\newcommand{\strings}{\mymacro{\alphabet^*}}

\newcommand{\bigo}{\mymacro{\mathcal{O}}}

\newcommand{\E}[0]{\mathop{\mathbb{E}}}

\DeclareMathOperator*{\argmax}{\mymacro{argmax}}

\newcommand{\entropy}{\mymacro{\mathrm{H}}}

\makeatletter
\newcommand{\pEns}{\@ifnextchar\bgroup\pEns@with\pEns@bare}
\newcommand{\pEns@bare}{\mymacro{\p_{\mathrm{ens}}}}
\newcommand{\pEns@with}[1]{\mymacro{\p_{\mathrm{ens}, #1}}}
\makeatother

\newcommand{\inputSet}{\mymacro{X}}
\newcommand{\outputSet}{\mymacro{Y}}
\newcommand{\funcSpace}{\mymacro{\outputSet^{\inputSet}}}
\newcommand{\dataset}{\mymacro{\mathcal{D}}}
\newcommand{\targetFunc}{\mymacro{f^\star}}
\newcommand{\symx}{\mymacro{x}}
\newcommand{\symy}{\mymacro{y}}
\newcommand{\equivClass}[1]{\mymacro{[#1]_{\sim}}}
\newcommand{\quotientSpace}{\mymacro{\funcSpace / {\sim}}}
\newcommand{\prior}{\mymacro{\p}}
\newcommand{\inputPrior}{\mymacro{\mu}}%
\newcommand{\programPrior}{\mymacro{\prior^0}}
\newcommand{\pushforward}{\mymacro{p_{\sim}}}
\newcommand{\proposalProg}{\mymacro{q}}%
\newcommand{\essThreshold}{\mymacro{\tau_{\mathrm{ESS}}}}%
\newcommand{\targetErr}{\mymacro{\epsilon}}%
\newcommand{\failProb}{\mymacro{\delta}}%
\newcommand{\evalSize}{\mymacro{N}}%
\newcommand{\numRounds}{\mymacro{T}}%
\newcommand{\round}{\mymacro{t}}%
\newcommand{\classvar}{\mymacro{c}}%
\newcommand{\lm}{\mymacro{\p_{\mathrm{LM}}}}
\newcommand{\promptOf}{\mymacro{\mathrm{prompt}}}%

\newcommand{\pushforwardHat}{\mymacro{{\hat \prior}_{\sim}}}

\newcommand{\nParticles}{\mymacro{M}}
\newcommand{\particleIdx}{\mymacro{m}}
\newcommand{\predictive}{\mymacro{p}}

\newcommand{\testSet}{\mymacro{S}}
\newcommand{\errPop}{\mymacro{\mathrm{err}_{\inputPrior}}}

\newcommand{\posterior}{\mymacro{p}}

\newcommand{\indicator}{\mathbbm{1}}

\newcommand{\prefixOp}[1]{\vec{{\normalcolor#1}}}
\newcommand{\pPrefix}[0]{\mymacro{\prefixOp{\p}_{\mathrm{LM}}}}

\newcommand{\eos}[0]{\ensuremath{\mymacro{\raisebox{-0.2pt}{$\mathsmaller{\square}$}}}\xspace}
\newcommand{\gI}[0]{\mymacro \gamma}%
\newcommand{\gIbar}[0]{\mymacro{\bar\gamma}}%
\newcommand{\roundAlt}{\mymacro{s}}%
\newcommand{\badEvent}{\mymacro{B}}%
\newcommand{\pacThreshold}{\mymacro{\tau}}%

\newcommand{\importanceWeight}[0]{\mymacro{w}}

\renewcommand{\hat}[1]{\widehat{#1}}

\usepackage{ellipsis}
\makeatletter
\renewcommand*{\mathellipsis}{%
  \mathinner{%
    \kern\ellipsisbeforegap%
    {\ldotp}\kern\ellipsisgap%
    {\ldotp}\kern\ellipsisgap%
    {\ldotp}\kern\ellipsisaftergap%
  }%
}
\renewcommand*{\dotsb@}{%
  \mathinner{%
    \kern\ellipsisbeforegap%
    {\cdotp}\kern\ellipsisgap%
    {\cdotp}\kern\ellipsisgap%
    {\cdotp}\kern\ellipsisaftergap%
  }%
}
\renewcommand*{\@cdots}{%
  \mathinner{%
    \kern\ellipsisbeforegap%
    {\cdotp}\kern\ellipsisgap%
    {\cdotp}\kern\ellipsisgap%
    {\cdotp}\kern\ellipsisaftergap%
  }%
}
\renewcommand*{\ellipsis@default}{%
  \ellipsis@before
  \kern\ellipsisbeforegap
  .\kern\ellipsisgap
  .\kern\ellipsisgap
  .\kern\ellipsisgap
  \ellipsis@after\relax}
\renewcommand*{\ellipsis@centered}{%
  \ellipsis@before
  \kern\ellipsisbeforegap
  .\kern\ellipsisgap
  .\kern\ellipsisgap
  .\kern\ellipsisaftergap
  \ellipsis@after\relax}
\AtBeginDocument{%
  \DeclareRobustCommand*{\dots}{%
    \ifmmode\@xp\mdots@\else\@xp\textellipsis\fi}}
\def\ellipsisgap{-.05em}
\def\ellipsisbeforegap{-.05em}
\def\ellipsisaftergap{.05em}
\makeatother

\colorlet{MacroColor}{black}\colorlet{MACROCOLOR}{black}%

\usepackage{thmtools, thm-restate}

\usepackage{wrapfig}
\usepackage{cleveref}%
\usepackage{multirow}%
\usepackage{siunitx}%
\usepackage{listings}
\definecolor{mintbg}{HTML}{f8f8f8}
\definecolor{mintframe}{HTML}{cccccc}
\lstdefinestyle{mintedlike}{
  backgroundcolor=\color{mintbg},
  basicstyle=\ttfamily,
  columns=flexible,
  keepspaces=true,
  breaklines=true,
  breakatwhitespace=false,
  frame=single,
  rulecolor=\color{mintframe},
  framesep=3mm,
  xleftmargin=20pt,
  xrightmargin=20pt,
  framexleftmargin=10pt,
  tabsize=4,
  showstringspaces=false,
  postbreak=\mbox{\textcolor{gray}{$\hookrightarrow$}\space}
}
\usepackage{upquote}
\usepackage[most]{tcolorbox}
\definecolor{promptph}{HTML}{27509B}
\lstdefinestyle{promptplain}{
  basicstyle=\ttfamily\footnotesize,
  columns=fixed,
  basewidth=0.5em,
  upquote=true,
  breaklines=true,
  breakatwhitespace=true,
  breakautoindent=true,
  breakindent=0pt,
  showstringspaces=false,
  moredelim=[s][\color{promptph}]{<}{>}
}
\newtcblisting{promptbox}[1]{
  enhanced, breakable, listing only,
  colback=mintbg, colframe=mintframe, boxrule=0.4pt, arc=1.5pt,
  colbacktitle=mintframe!45, coltitle=black, fonttitle=\small\bfseries, title={#1},
  left=6pt, right=6pt, top=3pt, bottom=3pt, toptitle=2pt, bottomtitle=2pt,
  before skip=8pt, after skip=10pt,
  listing options={style=promptplain}
}
\title{Interactive Program Induction with Language Model Priors}

\author{%
  Robin Shing Moon Chan\\
  ETH Zürich \\
  \texttt{robin.chan@inf.ethz.ch} \\
  \And
    Ryan Cotterell \\
  ETH Zürich\\
  \texttt{ryan.cotterell@inf.ethz.ch} \\
}

\begin{document}

\maketitle
\lhead{Preprint. Under review.}

\begin{abstract}
Program induction refers to the process in which a learner aims to recover a program from a few given input--output examples. 
This paradigm broadens programming access by letting users demonstrate the rule-based behavior they want to apply to unseen data, instead of having to implement it themselves.
However, recovering a program from finite input--output examples is fundamentally ambiguous, as a finite set of input--output examples is theoretically consistent with infinitely many programs whose output diverges on unseen inputs.
We view this problem from a Bayesian perspective, where, given a prior $\programPrior$ over programs, identifying the target program means concentrating posterior probability mass on it by observing additional input--output examples.
In the interactive program induction setting, the learner chooses maximally informative inputs and queries an oracle, such as the user, for their outputs.
We derive a greedy query-selection algorithm and show that it terminates after $\bigo(\tfrac{1}{\gI}\log 1/\programPrior(\targetFunc))$ informative oracle queries, and returns the target program $\targetFunc$ up to approximate equivalence when no other class outweighs it under the prior.
The algorithm uses a language model as the prior under a PAC-style relaxation of program equivalence, where the query-selection objective is estimated using importance sampling.
On the CodeARC benchmark, the algorithm improves accuracy over the LM only agent by up to $22$ percentage points while observing up to $10.5$ fewer input--output examples.

\medskip
{\centering
\raisebox{-0.25em}{\includegraphics[height=1em]{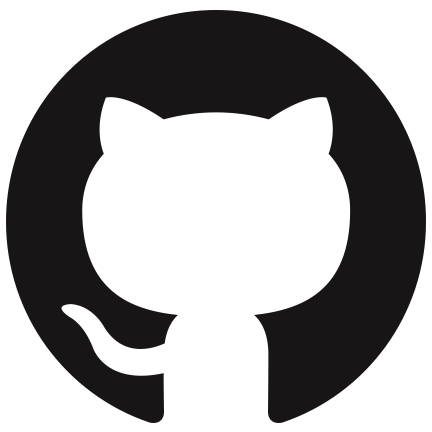}}
\url{https://github.com/chanr0/interactive-program-induction}\par}

\end{abstract}

\section{Introduction}\label{sec:introduction}

Many domain experts have a computer program clearly in mind that encodes certain rule-based behavior, but they cannot easily write it down themselves.
For instance, a spreadsheet user often knows how they would like the column of names to be reformatted, but cannot write the regex. Similarly, a linguist knows how a sentence in their native language should be segmented, but cannot program the tokenizer. \emph{Inductive programming}, also known as
programming by example
\citep{summers1977methodology,lieberman2001your,gulwani2012spreadsheet,gulwani2017program},
aims to bridge this gap by recovering the program from input--output examples.
The paradigm has been deployed at scale for specific applications, most prominently in Microsoft Excel's FlashFill
\citep{gulwani2011automating}.

However, recovering a target program from finite input--output examples is fundamentally ambiguous, since any finite set of examples is consistent with infinitely many programs that diverge on unseen data. Without an inductive bias, it is unclear which to choose \citep{solomonoff1964formal}. 
Prior work restricts the program space to a hand-designed domain-specific language (DSL) \citep{shapiro1983algorithmic, gulwani2017program}, at the cost of expressivity (cf. \cref{app:rel-work}). Within a DSL, the consistent programs can be enumerated, and the target one selected through human feedback, for instance by asking for further examples \citep{mayer2015user, ji2020question}, or through an oracle in place of a human \citep{solarlezama2008sketching, jha2010oracle}.
Large language models trained on code \citep{chen2021evaluatinglargelanguagemodels,austin2021programsynthesislargelanguage, li2022competition} may help to go beyond DSLs by providing distributions over programs in a general-purpose language (e.g., Python).
Still, when used as one-shot program inductors, they offer few mechanisms or guarantees for selecting the target among the many consistent programs.
\citet{li2024pbe} report a substantial gap to hand-engineered systems on programming-by-example, and \citet{wei2025codearc} report limited success of LLM agents doing autonomous program induction in Python with oracle feedback.\looseness=-1

In this work, we propose an algorithm for interactive program induction over a general-purpose programming language, and derive theoretical guarantees on the number of oracle queries it requires and on the accuracy of the recovered program. To this end, we treat the target program as the unknown of a Bayesian inference problem, with a language model prior $\programPrior$ over programs as inductive bias and a posterior that concentrates on $\targetFunc$ as input--output examples are observed, and we choose which inputs to query by sequential Bayesian experimental design \citep{lindley1956measure, mackay1992information, grand2026shoot, kobalczyk2025active}.
Using a notion of approximate equivalence between programs \citep[exact equivalence being undecidable,][]{rice1953classes} and an expected information gain criterion for query selection, we show that the algorithm terminates after $\bigo(\tfrac{1}{\gI}\log 1/\programPrior(\targetFunc))$ queries when every answer carries at least $\gI$ bits. \citet{dasgupta2004analysis} analyzed this greedy rule over finite hypothesis classes, and our analysis carries over to programs because we measure agreement on a finite sample of inputs. Conditional on returning the target's equivalence class, the returned program is approximately correct with high probability. Finally, we evaluate the practical implications on the CodeARC program induction dataset \citep{wei2025codearc} and show that the algorithm improves accuracy over the CodeARC LM agent while reducing the number of required oracle calls in most evaluated settings.

In summary, our contributions are as follows:
\begin{itemize}[leftmargin=*, topsep=2pt, itemsep=1pt]
  \item We formulate interactive program induction in a general-purpose language as Bayesian active learning. Since the posterior over programs is intractable and program equivalence is undecidable, we approximate the posterior, and with it the expected information gain, by importance sampling with a language model as prior and proposal (\cref{sec:importance}), and replace exact equivalence by agreement on a finite sample of inputs (\cref{sec:equivalence}).
  \item When every answer carries at least $\gI$ bits, we bound the number of queries by $\bigo(\tfrac{1}{\gI}\log 1/\programPrior(\targetFunc))$ per instance for any selection rule, and by the prior entropy over $\gI$ in expectation for the greedy rule. We characterize when a wrong class is returned and give a PAC bound on the error (\cref{sec:guarantees}).
  \item On CodeARC, our method improves accuracy over the LM-only agent on most evaluated language models, by up to $22$ percentage points with up to $10.5$ fewer input--output examples. Controls attribute the gains to the selection criterion and the importance weights (\cref{sec:experiments}, \cref{app:controls}). We find that the miscalibration of the language model prior limits the accuracy of the method (\cref{fig:prior-calibration}).
\end{itemize}

\section{Information-Theoretic Program Induction}

We formalize interactive program induction as the problem of recovering a \defn{target function} $\targetFunc$ from input--output observations that are revealed incrementally.
Let $\inputSet$ be a countable set of \defn{inputs} and $\outputSet$ a countable set of \defn{outputs}, and let the target $\targetFunc \colon \inputSet \to \outputSet$ be an unknown but fixed element of the candidate space $\funcSpace$.
We observe $\targetFunc$ only through its behavior on selected inputs. At each round $\round$ we choose an input $\symx^{(\round)} \in \inputSet$, the \defn{query}, and obtain its output $\symy^{(\round)} = \targetFunc(\symx^{(\round)})$ from the oracle.
Our goal is to return a function $\hat{\func}$ that closely approximates $\targetFunc$ after as few queries as possible.\looseness=-1

\subsection{Setup}\label{sec:assumptions}
A finite sample of input--output pairs does not pin down a target in $\funcSpace$, since infinitely many functions agree on any finite subset of $\inputSet$ and disagree elsewhere \citep{solomonoff1964formal}.
Recovering $\targetFunc$ therefore requires two pieces of additional structure, which together turn this ill-posed recovery problem into a well-posed Bayesian inference problem.
The first is a \defn{program prior} $\programPrior$ over $\funcSpace$, which encodes the inductive bias. Many functions are consistent with a finite observation set, and the prior determines which of them are a priori plausible.
The second is an \defn{input distribution} $\inputPrior$ over $\inputSet$, specifying the population of inputs on which the recovered function should agree with $\targetFunc$.
The program prior decides \emph{which} functions we prefer, and the input distribution decides \emph{where} agreement is measured.
We assume that programs are deterministic and that $\programPrior$ has countable support with $\programPrior(\targetFunc) > 0$. Whether a program halts on an input is undecidable, so evaluating whether it reproduces an observed output may not terminate. We therefore run every program with a time limit, and a program that raises or exceeds this limit on an input returns a distinguished error output, so every program defines a function on all of $\inputSet$.
In the setting most relevant to this work (program synthesis over strings, where $\inputSet = \outputSet = \strings$) the input distribution $\inputPrior$ has countably infinite support.
Comparing a candidate against $\targetFunc$ on every input in this support is infeasible, and even integrating against $\inputPrior$ to estimate disagreement requires structure we do not assume.
In the following, we derive the approximate notions of equivalence we use throughout.\looseness=-1

\subsection{Approximate Equivalence}\label{sec:equivalence}

Two facts about the interactive setting force us to relax exact program equivalence.
First, exact equivalence of two computable functions is undecidable \citep{rice1953classes}, so no algorithm can take two such functions and decide whether they agree on every input.
Second, two functions can be compared on any single input by running them, but the query budget is finite while $\inputSet$ is typically infinite, so agreement on every input cannot be certified by exhaustive testing.
We therefore want an equivalence relation that (i) measures disagreement under the population $\inputPrior$ and (ii) can be certified by inspecting only finitely many inputs.
The first relaxation accomplishes (i) by declaring $\func$ equivalent to $\targetFunc$ when their disagreement under $\inputPrior$ is small. For $\targetErr \in (0,1)$,
\begin{align}
  \func \sim_\targetErr \targetFunc
  \iff
  \errPop(\func) \le \targetErr,
  \qquad
  \errPop(\func)
    \defeq \Pr_{\symx \sim \inputPrior}[\func(\symx) \ne \targetFunc(\symx)].
  \label{eq:eps-equiv}
\end{align}
Checking $\sim_\targetErr$ requires integration against $\inputPrior$, which we cannot do directly.
The second relaxation makes equivalence checkable. We draw a random sample $\testSet = \{\symx_1, \ldots, \symx_\evalSize\}$ i.i.d. from $\inputPrior$ and replace agreement under $\inputPrior$ with agreement on $\testSet$,
\begin{align}
  \func \sim_\testSet \targetFunc
  \iff
  \func(\symx) = \targetFunc(\symx) \text{ for all } \symx \in \testSet.
  \label{eq:S-equiv}
\end{align}
Write $\equivClass{\func}$ for the $\sim_\testSet$-class of $\func$ and $\quotientSpace$ for the quotient, with $|\quotientSpace| \le |\outputSet|^\evalSize$.
Because $\testSet$ is finite, $\sim_\testSet$ can be verified on observed data. Because $\testSet$ is drawn from $\inputPrior$, agreement on $\testSet$ also controls $\inputPrior$-error with high probability over the draw.

\begin{restatable}[PAC Error Bound, cf.\ \citealt{blumer1987occam}]{proposition}{proppac}
\label{prop:pac-instance}
Let $\testSet = \{\symx_1, \ldots, \symx_\evalSize\}$ be drawn i.i.d. from $\inputPrior$.
With probability at least $1 - \failProb$ over the draw of $\testSet$, every
$\func \in \equivClass{\targetFunc}$ with $\programPrior(\func) > 0$ satisfies
\begin{align}
  \errPop(\func) \le \frac{\ln(1/\programPrior(\func)) + \ln(1/\failProb)}{\evalSize}.
  \label{eq:pac-instance}
\end{align}
\end{restatable}
\begin{proof}
    In \cref{app:proofs}.
\end{proof}

The bound in \cref{eq:pac-instance} depends on $\func$ through $\ln(1/\programPrior(\func))$. An $\func$ with more prior mass gets a smaller error bound from the same evaluation set, or reaches a given error bound with a smaller evaluation set.
Throughout the remainder of the paper, for notational brevity, $\sim$ and $\quotientSpace$ refer to $\sim_\testSet$ and $\funcSpace/\sim_\testSet$ for a fixed evaluation set $\testSet$ drawn once from $\inputPrior^\evalSize$.

\subsection{Bayesian Updating}
\label{sec:bayes}

With the equivalence relation in place, we set up the Bayesian inference machinery on which our query selection rule will operate.

\paragraph{Update rule.} Write $\dataset_0 \subset \inputSet \times \outputSet$ for the finite set of seed examples $(\symx, \targetFunc(\symx))$ given with the task \citep{wei2025codearc}, possibly empty, and $\dataset_\round = \dataset_0 \cup \{(\symx^{(\roundAlt)}, \symy^{(\roundAlt)})\}_{\roundAlt=1}^\round$ for the dataset after $\round$ oracle answers. The posterior over $\funcSpace$ given $\dataset_\round$ is
\begin{align}
\posterior(\func \mid \dataset_\round) = \frac{\programPrior(\func) \cdot \prod_{(\symx, \symy) \in \dataset_\round} \indicator\{\func(\symx) = \symy\}}{\sum_{\funcg \in \funcSpace} \programPrior(\funcg) \cdot \prod_{(\symx, \symy) \in \dataset_\round} \indicator\{\funcg(\symx) = \symy\}}.
\label{eq:posterior}
\end{align}
The posterior admits an equivalent recursive form,
\begin{subequations}\label{eq:posterior-recursive}
\begin{align}
  \posterior(\func \mid \dataset_\round) &= \frac{\posterior(\func \mid \dataset_{\round-1}) \cdot \indicator\{\func(\symx^{(\round)}) = \symy^{(\round)}\}}{\predictive(\symy^{(\round)} \mid \symx^{(\round)}, \dataset_{\round-1})},
  \label{eq:bayes-recursive}
  \\
  \predictive(\symy \mid \symx, \dataset_\round) &\defeq \sum_{\funcg \in \funcSpace} \posterior(\funcg \mid \dataset_\round) \cdot \indicator\{\funcg(\symx) = \symy\},
  \label{eq:posterior-predictive}
\end{align}
\end{subequations}
    where $\predictive(\symy \mid \symx, \dataset_\round)$ is the posterior predictive distribution of the output at an input $\symx$.
The recursive form \cref{eq:bayes-recursive} is the version we instantiate computationally in \cref{sec:importance}.

\paragraph{Pushforward onto $\quotientSpace$.}
By \cref{prop:pac-instance}, it suffices to identify the $\sim$-class of $\targetFunc$, since any $\func \in \equivClass{\targetFunc}$ has small $\errPop(\func)$ with high probability over $\testSet$.
The natural inference target is therefore the pushforward of the posterior on $\funcSpace$ onto the quotient,
\begin{align}
\pushforward(\equivClass{\func} \mid \dataset_\round) \defeq \sum_{\funcg \in \equivClass{\func}} \posterior(\funcg \mid \dataset_\round),
\label{eq:pushforward}
\end{align}
which assigns to each equivalence class the total posterior mass of its members.
\Cref{sec:greedy} selects queries to concentrate $\pushforward(\cdot \mid \dataset_\round)$ on a single class as quickly as possible.
 
\subsection{Greedy Information Gain Algorithm}\label{sec:greedy}

We now describe an algorithm that constructs $\dataset_1, \dataset_2,
\ldots$ by selecting a maximally informative query $\symx^{(\round)} \in \testSet$ at a time.
Queries are selected from $\testSet$ because that is where the classes are defined. All programs in a class return the same output on every $\symx \in \testSet$, so the answer to such a query eliminates whole classes and never splits one, and the posterior stays a distribution over the same classes. Any two distinct classes disagree on some $\symx \in \testSet$, so queries in $\testSet$ suffice to identify $\equivClass{\targetFunc}$.
Each query is chosen by the expected information gain criterion of Bayesian experimental design \citep{lindley1956measure, mackay1992information}, with the equivalence class of the target as the quantity of interest.
More precisely, at each round $\round = 1, 2, \ldots$:
\begin{enumerate}
  \item \textbf{Select.}
    Choose the input $\symx^{(\round)} \in \testSet$ that maximizes the
    expected information gain \citep{shannon1948mathematical, mackay1992information} with respect to
    $\pushforward(\cdot \mid \dataset_{\round-1})$,
    \begin{align}
      \symx^{(\round)}
        = \argmax_{\symx \in \testSet}
          \entropy\!\bigl[\pushforward(\cdot \mid \dataset_{\round-1})\bigr]
          - \sum_{\symy \in \outputSet}
            \predictive(\symy \mid \symx, \dataset_{\round-1})\,
            \entropy\!\bigl[
              \pushforward(\cdot \mid \dataset_{\round-1} \cup \{(\symx, \symy)\})
            \bigr],
      \label{eq:info-gain}
    \end{align}
    where $\entropy[\cdot]$ denotes the Shannon entropy and $\predictive(\cdot \mid \symx, \dataset_{\round-1})$ is the posterior predictive of \cref{eq:posterior-predictive}.

  \item \textbf{Query.}
    Observe $\symy^{(\round)} = \targetFunc(\symx^{(\round)})$ and set
    $\dataset_\round = \dataset_{\round-1} \cup \{(\symx^{(\round)}, \symy^{(\round)})\}$.

  \item \textbf{Update.}
    Compute $\posterior(\cdot \mid \dataset_\round)$ via
    \cref{eq:bayes-recursive} and the pushforward
    $\pushforward(\cdot \mid \dataset_\round)$ via \cref{eq:pushforward}.
\end{enumerate}
Terminate at the first round $\round \ge 0$ such that some class $\hat{\classvar}$ reaches majority pushforward mass, i.e., when
$\pushforward(\hat{\classvar} \mid \dataset_\round) > \tfrac{1}{2}$, and return any
$\hat\func \in \hat{\classvar}$. 
If $\hat\classvar = \equivClass{\targetFunc}$, the returned $\hat\func$ agrees with $\targetFunc$ on $\testSet$, and \cref{prop:pac-instance} gives $\errPop(\hat\func) \le (\ln(1/\programPrior(\hat\func)) + \ln(1/\failProb))/\evalSize$ except on an event of probability at most $\Pr[\hat\classvar \ne \equivClass{\targetFunc}] + \failProb$.

\section{Query Complexity and Correctness Guarantees}
\label{sec:guarantees}
We now derive a bound for the termination round $\numRounds$ of the greedy algorithm of \cref{sec:greedy} and characterize the returned program class $\hat \classvar$. Both are stated in terms of the entropy of the prior over classes and of the information a round is guaranteed to yield, which depends on the queries available in the evaluation set and on which of them the selection strategy picks.

\begin{restatable}[Expected query complexity and correctness]{theorem}{thmexpected}
\label{thm:expected}
Let $\targetFunc \sim \programPrior$, conditional on the seed examples $\dataset_0$, and let $ \entropy[\pushforward(\cdot \mid \dataset_0)] < \infty$ be the prior class entropy in bits. Write $\gIbar_\round \defeq \max_{\symx \in \testSet} -\log_2 \max_{\symy} \predictive(\symy \mid \symx, \dataset_{\round-1})$ for the most information an answer to a query in $\testSet$ is guaranteed to carry at round $\round$. The greedy strategy of \cref{sec:greedy} satisfies
\begin{enumerate}[label=(\roman*), leftmargin=*, topsep=2pt, itemsep=0pt]
    \item $\E_{\func^\star \sim \programPrior}\big[\sum_{\round=1}^{\numRounds} \gIbar_\round\big] \le \entropy[\pushforward(\cdot \mid \dataset_0)]$;
    \item $\numRounds \le \evalSize$ for every target;
    \item if $\gIbar_\round \ge \gI > 0$ for all $\round \le \numRounds$ and every $\targetFunc$ with $\programPrior(\targetFunc) > 0$, then \mbox{$\E_{\func^\star \sim \programPrior}[\numRounds] \le \entropy[\pushforward(\cdot \mid \dataset_0)]/\gI$}.
\end{enumerate}
At termination, $\Pr [\hat \classvar = \equivClass{\targetFunc} \mid \dataset_\numRounds] = \pushforward(\hat \classvar \mid \dataset_\numRounds) > \tfrac12$, i.e., for $\targetFunc \sim \programPrior$ the terminal mass is the exact probability that the returned program is $\sim_\testSet$-equivalent to $\targetFunc$. 
\end{restatable}
\begin{proof}
In \cref{app:proofs}.
\end{proof}

\begin{restatable}[Per-instance query complexity and correctness]{theorem}{thmworst}
\label{thm:worst}
Fix $\targetFunc$ with $\programPrior(\targetFunc) > 0$ and let $\entropy[\pushforward(\cdot \mid \dataset_0)] < \infty$. Write $\gI_\roundAlt \defeq -\log_2 \max_{\symy} \predictive(\symy \mid \symx^{(\roundAlt)}, \dataset_{\roundAlt-1})$ for the information the answer to the query of round $\roundAlt$ is guaranteed to carry. If $\pushforward(\equivClass{\targetFunc} \mid \dataset_0) \le \tfrac12$ and $\gI_\roundAlt \ge \gI > 0$ for all $\roundAlt \le \numRounds$, then any selection rule over $\testSet$ terminates after
\begin{align}
    \numRounds \le \left\lfloor \frac{1}{\gI}\, \log_2 \frac{1}{2 \pushforward(\equivClass{\targetFunc} \mid \dataset_0)} \right\rfloor + 1.
    \label{eq:T-worst}
\end{align}
Moreover, $\hat\classvar \ne \equivClass{\targetFunc}$ if and only if some class with prior mass strictly greater than $\pushforward(\equivClass{\targetFunc} \mid \dataset_0)$ is consistent with every query made. In particular, if no class outweighs $\equivClass{\targetFunc}$ under the prior, termination returns $\hat\classvar = \equivClass{\targetFunc}$, and under the hypotheses of \cref{eq:T-worst} it does so within that bound.
\end{restatable}
\begin{proof}
In \cref{app:proofs}.
\end{proof}

\Cref{thm:expected} bounds the number of queries in expectation over targets drawn from the prior. It is specific to the greedy rule, whose query has the largest expected information in $\testSet$ and therefore at least $\gIbar_\round$, the largest guaranteed information any query in $\testSet$ offers. \Cref{thm:worst} bounds the number of queries for a fixed target program. Its proof does not use the greedy rule, so the bound applies to every selection rule over $\testSet$ with the guaranteed information of that rule's own queries, and \cref{sec:experiments} empirically evaluates several such rules. In both, the rate of convergence is set by $\gIbar_\round$ over $\testSet$ and $\gI_\round$ at the selected query, which are determined by how the program prior spreads mass across classes and how finely $\testSet$, drawn from the input distribution, separates them.

\paragraph{Improvement over non-adaptive querying.}
Fix a target error $\targetErr$ and confidence $\failProb$. A non-adaptive strategy draws inputs at random from $\inputPrior$, has the oracle label all of them, and accepts a program that agrees with the target on every one. It needs $\Omega(\log(1/\failProb)/\targetErr)$ labeled inputs, since a program that disagrees with $\targetFunc$ on a set of $\inputPrior$-mass $\targetErr$ passes $m$ random inputs with probability $(1-\targetErr)^m \ge e^{-2m\targetErr}$ for $\targetErr \le \tfrac12$ \citep[see, e.g.,][]{shalev2014understanding}, and by \cref{prop:pac-instance} inputs of that order also suffice, up to the $\ln(1/\programPrior(\func))$ term. The greedy algorithm needs an evaluation set of the same size, but the oracle labels only the $\numRounds \le \evalSize$ inputs it queries. Since $\targetFunc$ lies in its own class, that class has prior mass at least $\programPrior(\targetFunc)$ whatever $\testSet$ is, so under the hypotheses of \cref{thm:worst} the algorithm asks at most $\lfloor \tfrac1{\gI} \log_2 \tfrac{1}{2\programPrior(\targetFunc)} \rfloor + 1$ queries, a number that does not depend on $\targetErr$. The non-adaptive cost grows with $1/\targetErr$, while the adaptive cost is set by how plausible the target is under the prior. The PAC certificate of \cref{prop:pac-instance} then holds for the returned program, conditional on the algorithm returning the target's class.

\paragraph{The role of $|\testSet|$.}\label{sec:pac-finite}
Combining \cref{prop:pac-instance} with \cref{thm:expected,thm:worst} gives a complete guarantee. The greedy algorithm terminates after at most $\lfloor \tfrac1{\gI} \log_2 \tfrac{1}{2\pushforward(\equivClass{\targetFunc} \mid \dataset_0)} \rfloor + 1$ queries when every answer carries at least $\gI$ bits, for a target drawn from the prior it returns the target's class with probability $\pushforward(\hat\classvar \mid \dataset_\numRounds) > \tfrac12$, and since \cref{prop:pac-instance} holds for every program in $\equivClass{\targetFunc}$ simultaneously, the returned program satisfies $\Pr[\errPop(\hat\func) > (\ln(1/\programPrior(\hat\func)) + \ln(1/\failProb))/\evalSize] \le \Pr[\hat\classvar \ne \equivClass{\targetFunc}] + \failProb$. The size of the evaluation set enters the two bounds in opposite directions. A larger $\testSet$ makes the error bound tighter, but it also refines $\sim$, so the target's class shrinks, its prior mass decreases, and the query bound grows. We evaluate the effect of $|\testSet|$ empirically in \Cref{sec:experiments}.

\section{Approximate Inference with a Language Model}\label{sec:importance}

The greedy algorithm of \cref{sec:greedy} requires the pushforward entropy $\entropy[\pushforward(\cdot \mid \dataset_\round)]$, a sum over equivalence classes weighted by posterior mass. The posterior is a distribution over the countably infinite support of the prior, so neither it nor this sum can be computed exactly. We approximate the posterior by a weighted set of programs sampled from a language model, and compute the \mbox{pushforward from the weights}.

\subsection{Language Model Proposal}\label{sec:lm-proposal}
\paragraph{Language model preliminaries.}
An \defn{alphabet} $\alphabet$ is a finite, non-empty set of symbols, $\strings$ the set of all strings $\str$ formed from $\alphabet$ (including the empty string), and $|\str|$ the length of $\str$. A \defn{language model} $\lm$ is a probability distribution over $\strings$. The \defn{prefix probability} $\pPrefix(\str) \defeq \sum_{\str' \in \strings} \lm(\str\,\str')$ is the probability that a string drawn from $\lm$ begins with $\str$. Whenever $\pPrefix(\str) > 0$, the \defn{conditional prefix probability} is
\begin{align}
\pPrefix(\str' \mid \str) \defeq
\begin{cases}
\lm(\str) / \pPrefix(\str) & \text{if } \str' = \eos, \\
\pPrefix(\str\,\str') / \pPrefix(\str) & \text{otherwise,}
\end{cases}
\label{eq:conditional-prefix}
\end{align}
where $\eos \notin \alphabet$ is a distinguished end-of-string delimiter. Iterating \cref{eq:conditional-prefix} yields the autoregressive factorization
\begin{align}
\lm(\str) = \pPrefix(\eos \mid \str) \prod_{\round=1}^{|\str|} \pPrefix(\sym_\round \mid \str_{<\round}),
\label{eq:factored-lm}
\end{align}
which computes $\lm(\str)$ from $|\str|+1$ next-token queries without enumerating $\strings$. The same definitions apply to a language model conditioned on a context string.

\paragraph{Proposal over programs.}
An LM is a distribution over strings, but our algorithm needs distributions over programs. We obtain them by prompting the LM with input--output examples and asking for a program consistent with them (\cref{app:prompts} gives the prompts), so that its distribution over completions is a distribution over source code. Strictly, this makes $\lm$ a distribution over \emph{implementations}, whereas a \emph{program} is an input--output behavior, and two distinct strings $\str_1 \ne \str_2$ may decode to the same function while $\lm(\str_1) \ne \lm(\str_2)$. The equivalence relation $\sim$ from \cref{sec:equivalence} and the pushforward $\pushforward$ from \cref{sec:bayes} collapse implementations that agree on $\testSet$ into a single class, so the algorithm operates at the level of programs, indifferent to syntactic differences between implementations, while sampling and density evaluation happen at the level of implementations, where \cref{eq:factored-lm} makes them tractable.

At each round $\round$ we encode the observed examples into a prompt \citep{chan2024affine}, a context string $\promptOf(\dataset_\round) \in \strings$ (\cref{app:prompts}), and condition the LM on it. Its completions are implementations $\str \in \strings$, each of which decodes to a function by execution. The proposal over implementations is
\begin{align}
  \proposalProg_\round(\str) \defeq \lm(\str \mid \promptOf(\dataset_\round)), \label{eq:q-proposal}
\end{align}
and we take the program prior to be the proposal before any query, $\programPrior \defeq \proposalProg_0$. The fixed objects $\programPrior$ and $\inputPrior$ define the guarantees of \cref{sec:guarantees}, while the proposal $\proposalProg_\round$ is an approximation mechanism whose quality affects practical efficiency but not the guarantees for the exact posterior.

\subsection{Importance Sampling Estimates}\label{sec:snis}
\paragraph{Importance weights.}
We draw $\nParticles$ implementations from $\proposalProg_\round$, write $\func_1, \ldots, \func_\nParticles$ for their decoded functions and $\programPrior(\func_\particleIdx)$, $\proposalProg_\round(\func_\particleIdx)$ for the densities of the sampled implementations, and assign self-normalizing importance weights
\begin{align}
  \importanceWeight_\particleIdx = \frac{\programPrior(\func_\particleIdx) \cdot \prod_{(\symx, \symy) \in \dataset_\round} \indicator\{\func_\particleIdx(\symx) = \symy\}}{\proposalProg_\round(\func_\particleIdx)}.
  \label{eq:importance-weight}
\end{align}
Programs inconsistent with the observed data $\dataset_\round$ receive zero weight. For consistent programs, $\programPrior(\func_\particleIdx)/\proposalProg_\round(\func_\particleIdx)$ corrects for the mismatch between prior and proposal. Because $\proposalProg_\round$ is already conditioned on $\dataset_\round$, it concentrates on consistent programs.
At round $0$ the weights reduce to the consistency indicator, since $\programPrior = \proposalProg_0$. Density ratios enter only when particles are resampled from a later $\proposalProg_\round$. Both densities are then the probability the LM assigns to the extracted code block under the respective prompt, without the reasoning that precedes it and without the sampling temperature. The weights are therefore an approximation to \cref{eq:importance-weight}, which we do because intermediate reasoning and a tuned sampling temperature improve the sampled programs (\cref{app:prompt-selection}). \Cref{app:controls} compares them to the exact weights of the sampled completions and to uniform weights. The approximate and exact weights perform on par, and both require significantly fewer input--output examples than uniform weights.\looseness=-1%

\paragraph{Pushforward mass.}
The pushforward mass on a class $\classvar$ is a posterior expectation, $\pushforward(\classvar \mid \dataset_\round) = \E_{\func \sim \posterior(\cdot \mid \dataset_\round)}[\indicator\{\func \in \classvar\}]$, which the self-normalized importance sampling (SNIS) estimator \citep{owen2013mc} replaces by the weighted share of the particles in the class,
\begin{align}
  \pushforwardHat(\classvar \mid \dataset_\round)
    = \frac{\sum_{\particleIdx:\, \func_\particleIdx \in \classvar} \importanceWeight_\particleIdx}{\sum_{\particleIdx=1}^\nParticles \importanceWeight_\particleIdx},
  \label{eq:mass-estimate}
\end{align}
where class membership is checked by comparing outputs on $\testSet$, as two programs share a $\sim$-class if and only if they agree on every $\symx \in \testSet$.

\paragraph{Entropy and information gain.}
Plugging \cref{eq:mass-estimate} into the definition of entropy gives the plug-in estimator
\begin{align}
  \hat\entropy[\pushforward(\cdot \mid \dataset_\round)]
    = -\sum_\classvar \pushforwardHat(\classvar \mid \dataset_\round) \log_2 \pushforwardHat(\classvar \mid \dataset_\round).
  \label{eq:entropy-estimate}
\end{align}
The information gain for a candidate query $\symx$ is estimated by partitioning the particles by predicted output $\func_\particleIdx(\symx)$, computing $\hat\entropy$ on each partition with renormalized weights, and weighting by the estimated posterior predictive
\begin{align}
  \hat\predictive(\symy \mid \symx, \dataset_\round)
    = \frac{\sum_{\particleIdx:\, \func_\particleIdx(\symx) = \symy} \importanceWeight_\particleIdx}{\sum_{\particleIdx=1}^\nParticles \importanceWeight_\particleIdx}.
  \label{eq:predictive-estimate}
\end{align}
The greedy step of \cref{sec:greedy} is a finite maximization over $\testSet$, with $\hat\entropy$ in place of $\entropy$ and $\hat\predictive$ in place of $\predictive$. \Cref{alg:greedy-ig} summarizes the resulting procedure, with an effective-sample-size test that redraws the particles from the current proposal when the weights degenerate.

\begin{algorithm}[t]
\caption{Greedy Information Gain Program Induction}
\label{alg:greedy-ig}
\begin{algorithmic}[1]
\Require Input distribution $\inputPrior$; program prior $\programPrior$; LM proposal
$\proposalProg_\round$; oracle for $\targetFunc$; evaluation size $\evalSize$; particle budget
$\nParticles$; ESS threshold $\essThreshold$.
\State Draw $\testSet = \{\symx_1, \ldots, \symx_\evalSize\}
       \overset{\mathrm{i.i.d.}}{\sim} \inputPrior$.
       \Comment{Fix evaluation set}
\State Sample $\func_1, \ldots, \func_\nParticles \sim \proposalProg_0$; compute $\importanceWeight_\particleIdx$ via
       \cref{eq:importance-weight}.
       \Comment{Initialize particles}
\State $\round \gets 0$; $\dataset_0 \gets$ seed examples.
\While{no class $\classvar$ has $\pushforwardHat(\classvar \mid \dataset_\round) > \tfrac{1}{2}$}
  \State $\round \gets \round + 1$.
  \State $\symx^{(\round)} \gets \argmax_{\symx \in \testSet}\,
         \hat\entropy[\pushforward(\cdot \mid \dataset_{\round-1})]
         - \sum_{\symy} \hat\predictive(\symy \mid \symx, \dataset_{\round-1})\,
         \hat\entropy[\pushforward(\cdot \mid \dataset_{\round-1} \cup
         \{(\symx, \symy)\})]$.
  \State Observe $\symy^{(\round)} \gets \targetFunc(\symx^{(\round)})$;
         $\dataset_\round \gets \dataset_{\round-1} \cup
         \{(\symx^{(\round)}, \symy^{(\round)})\}$.
         \Comment{Query oracle}
  \For{$\particleIdx = 1, \ldots, \nParticles$}
    \If{$\func_\particleIdx(\symx^{(\round)}) \ne \symy^{(\round)}$} $\importanceWeight_\particleIdx \gets 0$ \EndIf
  \EndFor
  \State Renormalize weights.
         \Comment{Eliminate inconsistent particles}
  \If{$\mathrm{ESS} = \bigl(\sum_\particleIdx \importanceWeight_\particleIdx\bigr)^2 / \sum_\particleIdx \importanceWeight_\particleIdx^2
      < \essThreshold$}
    \State Resample $\func_1, \ldots, \func_\nParticles \sim \proposalProg_\round$; recompute weights.
  \EndIf
\EndWhile
\State \Return $\hat\func$ in the majority class
       $\hat \classvar = \argmax_\classvar \pushforwardHat(\classvar \mid \dataset_\round)$.
\end{algorithmic}
\end{algorithm}

\paragraph{Approximation gaps.}
Under absolute continuity of the posterior with respect to $\proposalProg_\round$ (i.e., $\posterior(\func \mid \dataset_\round) > 0 \implies \proposalProg_\round(\func) > 0$), exact weights, and finitely many classes, the SNIS estimates $\pushforwardHat(\classvar \mid \dataset_\round)$ converge to $\pushforward(\classvar \mid \dataset_\round)$ as $\nParticles \to \infty$ \citep[e.g.,][]{owen2013mc}, and since $\hat\entropy$ is continuous in these estimates, $\hat\entropy[\pushforward(\cdot \mid \dataset_\round)] \to \entropy[\pushforward(\cdot \mid \dataset_\round)]$ by the continuous mapping theorem. The guarantees of \cref{sec:guarantees} are stated for the exact posterior, and for finite $\nParticles$ two things can go wrong. The particles may miss $\equivClass{\targetFunc}$ entirely, or a wrong class may exceed the threshold before enough of them land in $\equivClass{\targetFunc}$, which the miscalibration of the prior makes common (\cref{fig:prior-calibration}). \Cref{app:particles} varies the particle budget, and \cref{tab:failure-modes} and \cref{app:limitations} quantify how many runs terminate on a wrong class or without a consistent program.

\begin{table}[t]
  \centering
  \caption{Results across models and methods on the full anonymized CodeARC split. Entries are means over $5$ independent seeds with 95\% confidence intervals over seeds. Avg~I/O is the average number of input--output examples observed, including the seed invocations, out of a budget of $30$.}%
  \label{tab:results}

  \setlength{\aboverulesep}{0pt}
  \setlength{\belowrulesep}{0pt}

  \setlength{\tabcolsep}{5pt}
  \renewcommand{\arraystretch}{1.1}

  \newcommand{\ci}[2]{#1{\scriptscriptstyle\,\pm#2}}

  \footnotesize
  \begin{tabular}{@{} ll cc cc @{}}
    \toprule
    & & \multicolumn{4}{c}{\textit{Seed Invocations}} \\
    \cmidrule(lr){3-6}
    & & \multicolumn{2}{c}{1} & \multicolumn{2}{c}{10} \\
    \cmidrule(lr){3-4} \cmidrule(lr){5-6}
    {Model} & {Method}
      & {Avg I/O $\downarrow$} & {Acc. $\uparrow$}
      & {Avg I/O $\downarrow$} & {Acc. $\uparrow$} \\
    \midrule

    \multirow{2}{*}{Llama3.2-3B}
      & LM only (CodeARC) & $\ci{19.02}{0.44}$ & $\ci{.134}{.014}$ & $\ci{27.55}{0.16}$ & $\ci{.125}{.014}$ \\
      & \cellcolor{gray!10} \Cref{alg:greedy-ig} with EIG (ours)
        & \cellcolor{gray!10} $\mathbf{\ci{15.82}{0.36}}$
        & \cellcolor{gray!10} $\mathbf{\ci{.166}{.010}}$
        & \cellcolor{gray!10} $\mathbf{\ci{25.13}{0.14}}$
        & \cellcolor{gray!10} $\mathbf{\ci{.176}{.008}}$ \\
    \midrule

    \multirow{2}{*}{Llama3.1-8B}
      & LM only (CodeARC) & $\ci{16.60}{0.14}$ & $\ci{.129}{.015}$ & $\ci{26.86}{0.14}$ & $\ci{.142}{.009}$ \\
      & \cellcolor{gray!10} \Cref{alg:greedy-ig} with EIG (ours)
        & \cellcolor{gray!10} $\mathbf{\ci{12.89}{0.55}}$
        & \cellcolor{gray!10} $\mathbf{\ci{.212}{.011}}$
        & \cellcolor{gray!10} $\mathbf{\ci{22.77}{0.20}}$
        & \cellcolor{gray!10} $\mathbf{\ci{.273}{.015}}$ \\
    \midrule

    \multirow{2}{*}{Qwen2.5-7B}
      & LM only (CodeARC) & $\ci{19.54}{0.26}$ & $\ci{.148}{.010}$ & $\ci{27.88}{0.19}$ & $\ci{.156}{.015}$ \\
      & \cellcolor{gray!10} \Cref{alg:greedy-ig} with EIG (ours)
        & \cellcolor{gray!10} $\mathbf{\ci{9.01}{0.42}}$
        & \cellcolor{gray!10} $\mathbf{\ci{.229}{.008}}$
        & \cellcolor{gray!10} $\mathbf{\ci{20.92}{0.11}}$
        & \cellcolor{gray!10} $\mathbf{\ci{.322}{.017}}$ \\
    \midrule

    \multirow{2}{*}{Qwen2.5-32B}
      & LM only (CodeARC) & $\ci{11.13}{0.06}$ & $\ci{.250}{.011}$ & $\mathbf{\ci{18.75}{0.08}}$ & $\ci{.299}{.014}$ \\
      & \cellcolor{gray!10} \Cref{alg:greedy-ig} with EIG (ours)
        & \cellcolor{gray!10} $\mathbf{\ci{10.83}{0.32}}$
        & \cellcolor{gray!10} $\mathbf{\ci{.334}{.024}}$
        & \cellcolor{gray!10} $\ci{19.66}{0.20}$
        & \cellcolor{gray!10} $\mathbf{\ci{.415}{.010}}$ \\
    \midrule

    \multirow{2}{*}{Olmo3-7B}
      & LM only (CodeARC) & $\ci{13.34}{0.16}$ & $\ci{.064}{.007}$ & $\ci{21.42}{0.15}$ & $\ci{.113}{.016}$ \\
      & \cellcolor{gray!10} \Cref{alg:greedy-ig} with EIG (ours)
        & \cellcolor{gray!10} $\mathbf{\ci{5.90}{0.33}}$
        & \cellcolor{gray!10} $\mathbf{\ci{.240}{.014}}$
        & \cellcolor{gray!10} $\mathbf{\ci{19.62}{0.17}}$
        & \cellcolor{gray!10} $\mathbf{\ci{.332}{.013}}$ \\
    \midrule

    \multirow{2}{*}{Ministral-8B}
      & LM only (CodeARC) & $\ci{16.22}{0.12}$ & $\mathbf{\ci{.239}{.013}}$ & $\mathbf{\ci{22.91}{0.18}}$ & $\mathbf{\ci{.264}{.013}}$ \\
      & \cellcolor{gray!10} \Cref{alg:greedy-ig} with EIG (ours)
        & \cellcolor{gray!10} $\mathbf{\ci{13.54}{0.28}}$
        & \cellcolor{gray!10} $\ci{.170}{.010}$
        & \cellcolor{gray!10} $\ci{23.65}{0.24}$
        & \cellcolor{gray!10} $\ci{.226}{.012}$ \\
    \bottomrule
  \end{tabular}
\end{table}

\pagebreak
\section{Experiments}\label{sec:experiments}

We now evaluate the practical implications of our work on a programming induction benchmark in Python \citep[CodeARC;][]{wei2025codearc}. Its reference method, which we call the CodeARC LM agent, prompts a language model to autonomously choose the inputs to query,  propose an implementation, and terminate when it feels confident, with no posterior over programs and no criterion for the choice of inputs.
Our experiments aim to address four questions.
\begin{enumerate}[label=RQ\arabic*., leftmargin=*, labelwidth=2em]
    \item Does \cref{alg:greedy-ig} improve accuracy and oracle cost over the CodeARC LM agent?
    \item Does EIG ask queries with more guaranteed information than other selection criteria, and how much do the queries add over the prior alone?
    \item How does the choice of evaluation set size $\evalSize$ trade off convergence and correctness, as hypothesized in \cref{sec:pac-finite}?
    \item What are the failure modes of language model priors in program induction?
\end{enumerate}

\subsection{Setup}

\paragraph{Dataset.} We evaluate the proposed approach on the inductive inference benchmark CodeARC \citep{wei2025codearc}, whose $1114$ Python functions are drawn from HumanEval+, MBPP+, and APPS and range from $2$ to $74$ lines, from entry-level tasks to competition problems.
Similarly to our setup, the benchmark simulates an inductive inference scenario where a language modeling agent needs to find diverse Python functions from input--output pairs through membership queries (MQ), which query $\targetFunc$ on a chosen input $\symx$ to obtain $\targetFunc(\symx)$, and equivalence queries (EQ), which ask the oracle whether a candidate $\hat{\func}$ equals $\targetFunc$ and return a counterexample if it does not, under a fixed MQ and EQ invocation budget.\footnote{We note that our setup is equivalent to the CodeARC setup if the EQ budget is 1.}
We use the \emph{anonymized} split,
which replaces all function names with a generic identifier and so
suppresses any cues for the program prior.
Following \citet{wei2025codearc}, we provide a small number of
\emph{seed invocations} (initial input--output examples that we also use to
condition $\inputPrior$), evaluating both $1$ and $10$ seed invocations, and
set the query budget to $30$ input--output pairs, which includes the seed invocations. For consistency with their results, the CodeARC agent may additionally issue $2$ equivalence queries, while our algorithm makes none and submits its returned program once for evaluation.
We evaluate on the full split and report means over $5$ independent seed runs. The input distribution $\inputPrior$ fixes the inputs on which agreement with $\targetFunc$ is measured, and with it what equivalence means. It is part of the problem and independent of the seed invocations shown, so we condition it on all $10$ seed invocations of the benchmark and keep it fixed across settings (\cref{app:experimental-details}). For the models that overlap with \citet{wei2025codearc}, we reproduce their reported results within seed noise.

\paragraph{Models.} We evaluate our algorithm on six models of various sizes: Llama3.2-3B and Llama3.1-8B \citep{grattafiori2024llama3herdmodels}, Qwen2.5-7B and Qwen2.5-32B \citep{qwen2025qwen25technicalreport}, Olmo3-7B \citep{teamolmo2025olmo3}, and Ministral-8B \citep{mistral2024ministraux}. We use the instruction-tuned variant of every model.
We chose these models for their differing performance on the CodeARC benchmark.

\paragraph{Baselines and Evaluation.} 
To test whether our algorithm converges \textit{efficiently} on the target program, we compare it to the full CodeARC baselines in terms of average number of input--output examples observed and accuracy upon termination. CodeARC equivalence queries give the baseline strictly more oracle access than our method when the EQ budget exceeds 1, so CodeARC @ 2 EQ is a more powerful oracle setting. Following \citet{wei2025codearc}, we evaluate the correctness of the generated programs extrinsically using the differential testing tool \textsc{Mokav} \citep{etemadi2025moskav}. We fix \textsc{Mokav}'s test-generating language model to Qwen2.5-32B, the best-performing model, for all evaluated models and both methods, so that accuracies are comparable across models.\looseness=-1

\begin{figure}[!t]
    \centering
    \includegraphics[width=\linewidth]{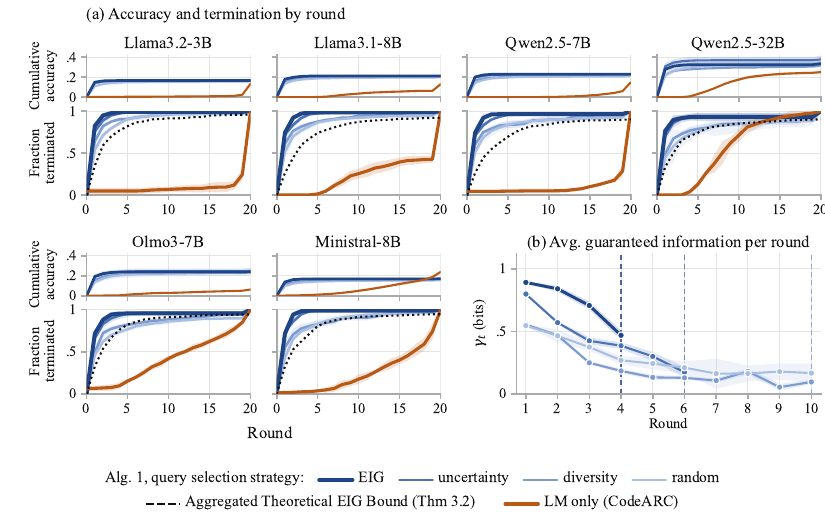}
    \caption{(a) Cumulative accuracy (top) and fraction of runs terminated (bottom) by round, for \cref{alg:greedy-ig} under four query selection strategies, the CodeARC agent (LM only), and the per-instance bound of \cref{thm:worst} aggregated over the EIG runs. The termination panels use the runs whose initial particles contain the target's class. A run is terminated once it reaches the total query budget. (b) The information the selected query is guaranteed to yield, $\gI_\round$, by round, averaged over the runs still querying and pooled across models. A dashed line marks the last round with at least $200$ runs of that strategy still querying, which is where the strategy has typically terminated. 95\% CIs over five seeds.}
    \label{fig:turn-results}
\end{figure}

\subsection{Results}\label{sec:results}

\paragraph{RQ1: \Cref{alg:greedy-ig} with EIG improves accuracy and query efficiency over the LM only agent.}
\cref{tab:results} shows that the proposed method improves over the LM only (CodeARC) agent in most of the evaluated settings in terms of the final accuracy (up to $+.22$) as well as oracle querying efficiency (up to $10.5$ fewer input--output examples). A gap to perfect accuracy remains, which we attribute to the models' general performance ceiling in proposing valid code matching the shown input--output pairs (cf. \citet{wei2025codearc}, and our failure mode analysis in \cref{tab:failure-modes}). Aside from this, the method commonly struggles with early termination due to early posterior concentration (RQ4), which is the case \cref{thm:worst} excludes, a heavier class that stays consistent with every answer. Our ablation in \cref{fig:prompt-temperature} shows that the choice of proposal prompt and temperature reduces this.\looseness=-1
\paragraph{RQ2: EIG asks the most informative queries and terminates in the fewest rounds, at an accuracy within noise of the other criteria.}
The termination panels of \cref{fig:turn-results} condition on runs whose initial particles contain the target's class, for all strategies alike. Compared to the other selection criteria within the proposed algorithm, EIG selects the most informative queries, $.70$ to $.99$ guaranteed bits per query against $.31$ to $.52$ for diversity and random selection, terminates in the fewest rounds, and reaches an accuracy within the seed intervals of the best criterion for every model (\cref{fig:turn-results}, \cref{tab:strategies}). The queries yield a significant improvement over two controls, prior majority, which returns a program from the majority class of the initial particles without any query, and the CodeARC agent, which selects its queries heuristically (\cref{tab:results,tab:strategies}). \Cref{tab:bound} reports the bound of \cref{thm:worst} on the EIG runs, which is attained in $.37$ of them. \Cref{app:controls} further compares against uniform weights, minimax selection, and a compute-matched propose-and-verify baseline.

\begin{figure}[t]
    \centering\setlength{\abovecaptionskip}{3pt}
    \begin{minipage}[t]{0.56\textwidth}
        \centering
        \includegraphics[height=1.45in,trim=7 9 7 10,clip]{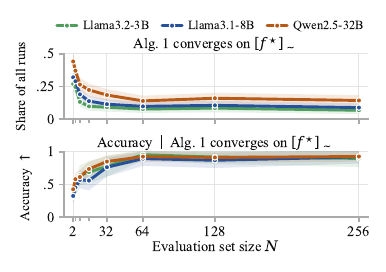}
        \caption{$\evalSize$ ablation on three models. Top: share of runs converging on $\equivClass{\targetFunc}$. Bottom: accuracy given convergence. 95\% CIs over runs, three seeds.}
        \label{fig:nt-tradeoff}
    \end{minipage}\hfill
    \begin{minipage}[t]{0.4\textwidth}
        \centering
        \includegraphics[height=1.45in]{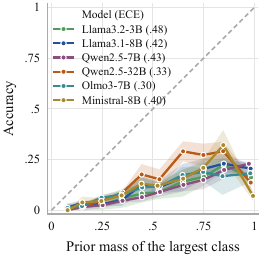}
        \caption{Calibration of the prior majority class,  with ECE in the legend. 95\% CIs over five seeds.}
        \label{fig:prior-calibration}
    \end{minipage}
\end{figure}

\paragraph{RQ3: Larger $\testSet$ makes convergence on the target's class rarer but more reliable.}
\Cref{fig:nt-tradeoff} presents an ablation of the evaluation set size $\evalSize$ from 2 to 256 for three models. As predicted in \Cref{sec:pac-finite}, a smaller $\evalSize$ makes $\equivClass{\targetFunc}$ coarse, since equivalence then requires agreement on only a few inputs. The posterior concentrates early on such a coarse class, and the program returned from it is often rejected by the oracle despite agreeing with $\targetFunc$ on $\testSet$. As $\testSet$ grows, convergence on $\equivClass{\targetFunc}$ becomes rarer but a reliable indicator of correctness. The rising conditional accuracy instantiates \cref{prop:pac-instance}, since agreement with $\targetFunc$ on more sampled inputs means fewer disagreements on unseen ones.

\paragraph{RQ4: The language model prior is miscalibrated.}
\Cref{fig:prior-calibration} is a calibration plot of the prior majority class against the CodeARC targets.\footnote{The CodeARC targets may not coincide with the programs humans would consider plausible, against which the prior may be calibrated differently.} For every model, the majority class is the target's class far less often than its prior mass suggests. The prior is thus confident early, while Bayesian inductive inference relies on the prior to keep alternatives plausible until the data rule them out. By \cref{thm:worst}, a run converges on a wrong program if the prior strongly favors a wrong class, which this miscalibration makes common (\cref{tab:failure-modes}).\looseness=-1

\section{Conclusion}
We presented a Bayesian active-learning approach for efficient program induction with oracle feedback that uses a language model as both the program prior and the importance-sampling proposal. We proved that the greedy information-gain algorithm terminates after $\bigo(\tfrac{1}{\gI}\log 1/\programPrior(\targetFunc))$ oracle invocations when every answer carries at least $\gI$ bits, and introduced an equivalence relation that yields a PAC certificate on population error. Empirically, the algorithm improves accuracy and reduces oracle calls over the CodeARC agent in most evaluated settings, and its queries improve over returning the prior's majority class. We see this work as a step toward principled human-in-the-loop program induction, where the same framework can accommodate richer oracle models, such as noisy users, and structured priors.\looseness=-1

\bibliography{main}
\bibliographystyle{iclr2027_conference}

\appendix

\section{Related Work}\label{app:rel-work}

\paragraph{Interactive and oracle-guided program synthesis.}
Early works  in oracle-guided program synthesis include CEGIS
\citep{solarlezama2008sketching}, and is formalized in works by \citet{jha2010oracle} and \citet{jha2017theory}, where an automated verifier supplies counterexamples in place of a human. Interactive disambiguation with human oracles was deployed in FlashFill by \citet{mayer2015user} and studied theoretically by \citet{ji2020question}, who select questions by minimax over programs sampled from a PCFG posterior over a version space. \citet{tiwari2020information} select inputs that maximize the entropy of the outputs of sampled programs, weighted by a ranking heuristic. All of these operate over a finite DSL and use heuristic or symbolic distinguishability criteria. We work over a countably infinite program space and derive query-complexity bounds from the program prior.

\paragraph{Active learning for program synthesis.}
\citet{barnaby2025active} discuss active learning in the neurosymbolic
setting, where neural components can mispredict and cause convergence
failures. Their constrained conformal evaluation is complementary to
our work, which assumes a symbolic target and bounds query complexity
information-theoretically. Our query-selection criterion is an instance
of Bayesian experimental design \citep{lindley1956measure, mackay1992information}, applied sequentially as in active learning \citep{settles2012active}.
\citet{dasgupta2004analysis} showed that greedy information-gain maximization over a \emph{finite} class uses at most $4\ln(1/\min_h \programPrior(h))$ times the expected number of queries of the optimal strategy. We bound the number of queries of the greedy rule over a countable program space by measuring agreement on a finite sample of inputs, and replace enumeration with importance sampling under an LM proposal. Our query bounds depend on the size of the evaluation set, because a larger set separates more programs into distinct classes. A finite hypothesis class has no such dependence. \citet{golovin2010near}, \citet{golovin2011adaptive}, and \citet{nowak2011geometry} show that greedy rules over finite classes use nearly as few queries as the best possible adaptive plan of queries, which is intractable to compute. We bound the number of queries directly and make no such theoretical comparison.\looseness=-1

\paragraph{Bayesian program induction.}
Bayesian priors over programs underlie a long line of work.
\citet{lake2015human} used hand-designed generative models over strokes
for few-shot concept learning, \citet{ellis2021dreamcoder} bootstrapped
reusable libraries via wake-sleep inference, \citet{ellis2022synthesizing}
applied the same framework to linguistic theory synthesis, and
\citet{saad2019bayesian} sampled probabilistic programs from PCFG
posteriors. These systems trade expressivity for tractability through a
hand-crafted DSL or generative model. We keep the Bayesian perspective
but substitute a language model for the hand-crafted prior, using
importance sampling to make inference tractable over an unrestricted
program space. \citet{singhal2026llm} use a language model to propose programs for empirical risk minimization on a fixed sample and give generalization guarantees. We select the examples interactively.\looseness=-1

\paragraph{Information-seeking questions with language models.}
Several works select questions by expected information gain over hypotheses sampled from a language model, for natural-language rules on a Zendo-style task \citep{piriyakulkij2024doing}, task disambiguation including code generation \citep{kobalczyk2025active}, collaborative search games \citep{grand2026shoot}, and natural-language database querying \citep{chan2026pleasqlarify}. \citet{lee2025program} select inputs by maximin over the outputs of language-model-sampled programs, with a language model predicting outputs in place of the oracle. \citet{kobalczyk2025active} and \citet{lee2025program} treat sampled hypotheses as equally likely. We weight them by importance sampling under an explicit program prior. We apply this idea to program induction with an input--output oracle and derive query-complexity and correctness guarantees under an explicit program prior.\looseness=-1

\paragraph{LLMs for program synthesis.}
LLMs achieve strong results on natural-language-specified code
generation \citep{chen2021evaluatinglargelanguagemodels,austin2021programsynthesislargelanguage,li2022competition}
but struggle with programming by example without domain-specific
fine-tuning \citep{li2024pbe}. HySynth \citep{barke2024hysynthcontextfreellmapproximation}
extracts a task-specific PCFG from LLM samples to guide bottom-up
enumerative search, and \citet{li2024guiding} pursue a similar direction using syntactic guidance. Both treat the LM as a one-shot prior over a fixed DSL, without user interaction or posterior updating. Grouping sampled programs by their outputs on a set of inputs is used to select a submission in AlphaCode \citep{li2022competition}, CodeT \citep{chen2023codet}, and MBR-exec \citep{shi2022natural}, and our equivalence relation $\sim_\testSet$ is the same construction. \citet{wei2025codearc} introduced CodeARC, an interactive benchmark where agents iteratively query a hidden oracle and input selection is left to the agent's heuristics. We use the LM as a prior and importance sampling proposal within a Bayesian framework and choose oracle membership queries to maximize expected information gain.

\section{Limitations}\label{app:limitations}

\paragraph{Computational cost.}
Each round of the algorithm requires potentially drawing $\nParticles$ program samples from the LM
proposal $\proposalProg_\round$, evaluating each on the inputs in $\testSet$, and recomputing
importance weights and entropy estimates over the candidate pool.
Larger evaluation sets, larger particle budgets, and longer programs all
compound this cost, and the per-round latency is dominated by LM inference.
\cref{tab:cost} reports the cost per problem. \cref{alg:greedy-ig} makes $32$ to $78$ LM calls and samples about $110$ candidate programs, each of which it executes on $\testSet$ and on every query, against $5$ to $10$ calls for the agent. Its wall-clock time per problem is nevertheless below the agent's in our setup, since the agent's calls are long generations and it keeps querying for most of its budget. The query-efficiency gains we report are savings in oracle effort, and the compute cost is in LM calls and program executions. Finally, the importance weights at resampling require the LM's token log-probabilities of a program under two prompts, which closed APIs may not expose.

\begin{table}[!t]
  \centering
  \caption{Computational cost per problem with $1$ seed invocation on the full anonymized CodeARC split, means over runs. LM calls are generation requests to the language model. Programs are the candidate programs sampled by \cref{alg:greedy-ig} and the implementations submitted by the agent. Wall-clock times were measured on different clusters under concurrent load and are indicative.}
  \label{tab:cost}
  \setlength{\aboverulesep}{0pt}
  \setlength{\belowrulesep}{0pt}
  \setlength{\tabcolsep}{4pt}
  \renewcommand{\arraystretch}{1.0}
  \scriptsize
  \begin{tabular}{@{} ll ccc @{}}
    \toprule
    {Model} & {Method} & {Wall-clock (s) $\downarrow$} & {LM calls $\downarrow$} & {Programs} \\
    \midrule
    \multirow{2}{*}{Llama3.2 3B}
      & \cellcolor{gray!10} \Cref{alg:greedy-ig} with EIG (ours) & \cellcolor{gray!10} $102$ & \cellcolor{gray!10} $33.1$ & \cellcolor{gray!10} $112.9$ \\
      & LM only (CodeARC) & $289$ & $4.7$ & $2.6$ \\
    \midrule
    \multirow{2}{*}{Llama3.1 8B}
      & \cellcolor{gray!10} \Cref{alg:greedy-ig} with EIG (ours) & \cellcolor{gray!10} $184$ & \cellcolor{gray!10} $49.8$ & \cellcolor{gray!10} $110.9$ \\
      & LM only (CodeARC) & $330$ & $7.1$ & $2.8$ \\
    \midrule
    \multirow{2}{*}{Qwen2.5 7B}
      & \cellcolor{gray!10} \Cref{alg:greedy-ig} with EIG (ours) & \cellcolor{gray!10} $142$ & \cellcolor{gray!10} $72.2$ & \cellcolor{gray!10} $106.4$ \\
      & LM only (CodeARC) & $299$ & $7.4$ & $2.5$ \\
    \midrule
    \multirow{2}{*}{Qwen2.5 32B}
      & \cellcolor{gray!10} \Cref{alg:greedy-ig} with EIG (ours) & \cellcolor{gray!10} $244$ & \cellcolor{gray!10} $77.8$ & \cellcolor{gray!10} $105.1$ \\
      & LM only (CodeARC) & $991$ & $9.3$ & $2.5$ \\
    \midrule
    \multirow{2}{*}{Olmo3 7B}
      & \cellcolor{gray!10} \Cref{alg:greedy-ig} with EIG (ours) & \cellcolor{gray!10} $258$ & \cellcolor{gray!10} $72.0$ & \cellcolor{gray!10} $108.6$ \\
      & LM only (CodeARC) & $350$ & $10.1$ & $0.9$ \\
    \midrule
    \multirow{2}{*}{Ministral 8B}
      & \cellcolor{gray!10} \Cref{alg:greedy-ig} with EIG (ours) & \cellcolor{gray!10} $182$ & \cellcolor{gray!10} $32.1$ & \cellcolor{gray!10} $115.5$ \\
      & LM only (CodeARC) & $410$ & $6.6$ & $2.6$ \\
    \bottomrule
  \end{tabular}
\end{table}

\paragraph{Dependence on the input distribution.}
\Cref{prop:pac-instance} bounds the error under the input distribution $\inputPrior$, which we instantiate by prompting an LM with the seed examples. A single seed invocation suffices for the gains of \cref{tab:results}, and rebuilding the evaluation set from the single visible seed example changes accuracy by at most $.02$ on five of six models (\cref{app:experimental-details}). The bound does not guarantee generalization beyond $\inputPrior$. A differential tester searches for a counterexample and may find one on an input that $\inputPrior$ rarely produces, so the accuracies we report also measure how well agreement under $\inputPrior$ generalizes. A prompt that describes the intended inputs more closely improves this generalization.

\paragraph{Gap between theoretical and empirical correctness.}
\Cref{thm:expected,thm:worst} hold for the exact posterior. The algorithm works with a finite particle approximation of it, and with priors that are sometimes concentrated on a wrong class. Both account for the $.34$ to $.63$ of runs in \cref{tab:failure-modes} that converge on a wrong program, since a wrong class can exceed the threshold before enough particles land in the target's class, or hold most of the prior mass, as the miscalibrated prior of \cref{fig:prior-calibration} often does. The remaining errors are due to the models' capability. In $.13$ to $.49$ of the runs no sampled program is consistent with the observed examples, and the CodeARC agent built on the same models reaches $.06$ to $.30$ (\cref{tab:results}). Stronger models narrow this gap, with Qwen2.5-32B reaching the highest accuracy in every setting.\looseness=-1

\section{Deferred Proofs}\label{app:proofs}

\subsection{PAC Bound}
\proppac*
\begin{proof}
Fix $\func$ with $\programPrior(\func) > 0$. Since $\testSet$ consists of $\evalSize$ i.i.d.\ draws from $\inputPrior$, the probability that $\func$ agrees with $\targetFunc$ on all of $\testSet$ is
\begin{align}
    \Pr_{\testSet \sim \inputPrior^\evalSize}\big[\func \in \equivClass{\targetFunc}\big] = (1 - \errPop(\func))^\evalSize \le \exp(-\evalSize \cdot \errPop(\func)),
    \label{eq:agree-prob}
\end{align}
where we used $1 - u \le e^{-u}$. Let $\pacThreshold(\func) \defeq (\ln(1/\programPrior(\func)) + \ln(1/\failProb))/\evalSize$ be the right-hand side of \cref{eq:pac-instance}, and let $\badEvent_\func$ be the event that $\func \in \equivClass{\targetFunc}$ and $\errPop(\func) > \pacThreshold(\func)$. The event $\badEvent_\func$ is empty if $\errPop(\func) \le \pacThreshold(\func)$. Otherwise \cref{eq:agree-prob} gives $\Pr_{\testSet \sim \inputPrior^\evalSize}[\badEvent_\func] \le \exp(-\evalSize \cdot \pacThreshold(\func)) = \failProb \cdot \programPrior(\func)$. The probability that any of these events occurs is at most the sum of their probabilities, so
\begin{align}
    \Pr_{\testSet \sim \inputPrior^\evalSize}\Big[\bigcup_{\func \colon \programPrior(\func) > 0} \badEvent_\func\Big] \le \sum_{\func \colon \programPrior(\func) > 0} \failProb \cdot \programPrior(\func) \le \failProb.
\end{align}
Assigning each program the failure probability $\failProb \cdot \programPrior(\func)$ instead of $\failProb/|\funcSpace|$ replaces $\ln|\funcSpace|$ by $\ln(1/\programPrior(\func))$ and extends the bound to countably infinite program spaces.
On the complement of this event, every $\func \in \equivClass{\targetFunc}$ with $\programPrior(\func) > 0$ satisfies \cref{eq:pac-instance}.
\end{proof}

\subsection{Expected Query Complexity and Correctness}
Throughout, $\testSet$ is fixed, and for a class $\classvar$ and $\symx \in \testSet$ we write $\classvar(\symx)$ for the common output of the programs in $\classvar$. A class is \emph{consistent} with $\dataset_\round$ if $\classvar(\symx^{(\roundAlt)}) = \symy^{(\roundAlt)}$ for all $\roundAlt \le \round$, and $\equivClass{\targetFunc}$ is consistent with every $\dataset_\round$. Ties in \cref{eq:info-gain} are broken by a fixed rule.

\begin{lemma}[Mass recursion]\label{lem:recursion}
For every class $\classvar$ and $1 \le \round \le \numRounds$,
\begin{align}
    \pushforward(\classvar \mid \dataset_\round) &= \frac{\pushforward(\classvar \mid \dataset_{\round-1}) \cdot \indicator\{\classvar(\symx^{(\round)}) = \symy^{(\round)}\}}{\predictive(\symy^{(\round)} \mid \symx^{(\round)}, \dataset_{\round-1})},
    \label{eq:class-recursion} \\
    \log_2 \frac{1}{\pushforward(\equivClass{\targetFunc} \mid \dataset_0)} &= \sum_{\roundAlt=1}^{\round} \log_2 \frac{1}{\predictive(\symy^{(\roundAlt)} \mid \symx^{(\roundAlt)}, \dataset_{\roundAlt-1})} + \log_2 \frac{1}{\pushforward(\equivClass{\targetFunc} \mid \dataset_\round)}.
    \label{eq:bits}
\end{align}
\end{lemma}
\begin{proof}
All programs in a class agree on $\symx^{(\round)} \in \testSet$, so summing \cref{eq:bayes-recursive} over $\funcg \in \classvar$ gives \cref{eq:class-recursion}. For $\classvar = \equivClass{\targetFunc}$ the indicator equals one, and unrolling over $\roundAlt \le \round$ and taking logarithms gives \cref{eq:bits}.
\end{proof}

\begin{lemma}[Greedy selection]\label{lem:greedy}
Let $\entropy[\pushforward(\cdot \mid \dataset_0)] < \infty$. For every $\round \le \numRounds$ and $\symx \in \testSet$, the objective in \cref{eq:info-gain} equals $\entropy[\predictive(\cdot \mid \symx, \dataset_{\round-1})]$. Moreover, $\gI_\round > 0$ for all $\round \le \numRounds$, no input is queried twice, and $\numRounds \le \evalSize$.
\end{lemma}
\begin{proof}
Let $\classvar \sim \pushforward(\cdot \mid \dataset_{\round-1})$, whose entropy is at most $\entropy[\pushforward(\cdot \mid \dataset_0)] / \pushforward(\equivClass{\targetFunc} \mid \dataset_0) < \infty$. The answer $\symy = \classvar(\symx)$ is a function of $\classvar$ with distribution $\predictive(\cdot \mid \symx, \dataset_{\round-1})$, and by \cref{lem:recursion} the distribution of $\classvar$ given $\symy$ is $\pushforward(\cdot \mid \dataset_{\round-1} \cup \{(\symx, \symy)\})$. The chain rule for entropy gives
\begin{align}
    \entropy[\pushforward(\cdot \mid \dataset_{\round-1})] = \entropy[\predictive(\cdot \mid \symx, \dataset_{\round-1})] + \sum_{\symy \in \outputSet} \predictive(\symy \mid \symx, \dataset_{\round-1})\, \entropy[\pushforward(\cdot \mid \dataset_{\round-1} \cup \{(\symx, \symy)\})],
\end{align}
which is the first claim. For $\round \le \numRounds$, no class has mass above $\tfrac12$ under $\dataset_{\round-1}$, so two classes have positive mass and differ on some $\symx' \in \testSet$. Then $\entropy[\predictive(\cdot \mid \symx^{(\round)}, \dataset_{\round-1})] \ge \entropy[\predictive(\cdot \mid \symx', \dataset_{\round-1})] > 0$, the predictive at $\symx^{(\round)}$ is not a point mass, and $\gI_\round > 0$. The predictive at a previously queried input is a point mass, so no input is queried twice. Once all of $\testSet$ is queried, $\equivClass{\targetFunc}$ is the only consistent class, and thus $\numRounds \le \evalSize$.
\end{proof}

\begin{lemma}[Posterior as conditional law]\label{lem:conditional}
Let $\targetFunc \sim \programPrior$, conditional on the seed examples $\dataset_0$. For every class $\classvar$ and every $\round \le \numRounds$, we have $\Pr[\equivClass{\targetFunc} = \classvar \mid \dataset_\round] = \pushforward(\classvar \mid \dataset_\round)$.
\end{lemma}
\begin{proof}
The queries are a deterministic function of the answers received so far. A target in class $\classvar$ therefore generates $\dataset_\round$ if and only if $\classvar$ is consistent with it, that is, $\Pr[\dataset_\round \mid \equivClass{\targetFunc} = \classvar] = \indicator\{\classvar \text{ consistent with } \dataset_\round\}$. Bayes' rule with $\Pr[\equivClass{\targetFunc} = \classvar] = \pushforward(\classvar \mid \dataset_0)$ gives the claim.
\end{proof}

\thmexpected*
\begin{proof}
Claim (ii) follows from \cref{lem:greedy}. For (i), dropping the last term of \cref{eq:bits} at $\round = \numRounds$, which is non-negative, gives $\sum_{\round=1}^{\numRounds} -\log_2 \predictive(\symy^{(\round)} \mid \symx^{(\round)}, \dataset_{\round-1}) \le -\log_2 \pushforward(\equivClass{\targetFunc} \mid \dataset_0)$. We take expectations over $\targetFunc \sim \programPrior$. The right-hand side becomes $\entropy[\pushforward(\cdot \mid \dataset_0)]$ since $\Pr[\equivClass{\targetFunc} = \classvar] = \pushforward(\classvar \mid \dataset_0)$. On the left-hand side, the event $\{\numRounds \ge \round\}$ is determined by $\dataset_{\round-1}$, and by \cref{lem:conditional} the answer $\symy^{(\round)}$ given $\dataset_{\round-1}$ is distributed as $\predictive(\cdot \mid \symx^{(\round)}, \dataset_{\round-1})$, so that the conditional expectation of the $\round$-th term given $\dataset_{\round-1}$ is $\entropy[\predictive(\cdot \mid \symx^{(\round)}, \dataset_{\round-1})]$. Summing over $\round \le \evalSize$ yields
\begin{align}
    \E_{\targetFunc \sim \programPrior}\Big[\sum_{\round=1}^{\numRounds} \entropy[\predictive(\cdot \mid \symx^{(\round)}, \dataset_{\round-1})]\Big] \le \entropy[\pushforward(\cdot \mid \dataset_0)].
\end{align}
By \cref{lem:greedy} and because the Shannon entropy of a distribution is at least the negative logarithm of its largest probability, $\entropy[\predictive(\cdot \mid \symx^{(\round)}, \dataset_{\round-1})] = \max_{\symx \in \testSet} \entropy[\predictive(\cdot \mid \symx, \dataset_{\round-1})] \ge \gIbar_\round$, which proves (i). Keeping the last term of \cref{eq:bits} and applying \cref{lem:conditional} once more shows that the left-hand side equals $\entropy[\pushforward(\cdot \mid \dataset_0)] - \E[\entropy[\pushforward(\cdot \mid \dataset_\numRounds)]]$, so the only inequalities behind (i) are $\entropy[\pushforward(\cdot \mid \dataset_\numRounds)] \ge 0$ and this lower bound on the Shannon entropy. For (iii), the hypothesis gives $\gI \numRounds \le \sum_{\round=1}^{\numRounds} \gIbar_\round$ for every $\targetFunc$ with $\programPrior(\targetFunc) > 0$, and taking expectations and applying (i) gives the bound. Finally, the returned class $\hat\classvar$ is determined by $\dataset_\numRounds$, so \cref{lem:conditional} gives $\Pr[\hat\classvar = \equivClass{\targetFunc} \mid \dataset_\numRounds] = \pushforward(\hat\classvar \mid \dataset_\numRounds)$, which exceeds $\tfrac12$ by the stopping rule.
\end{proof}

\subsection{Per-Instance Query Complexity and Correctness}
\begin{lemma}[Mass growth]\label{lem:mass-growth}
Fix $\targetFunc$ with $\programPrior(\targetFunc) > 0$. Under any rule that selects queries from $\testSet$, for every $\round \le \numRounds$,
\begin{align}
    \pushforward\big(\equivClass{\targetFunc} \mid \dataset_\round\big) \ge \pushforward(\equivClass{\targetFunc} \mid \dataset_0) \cdot 2^{\sum_{\roundAlt=1}^{\round} \gI_\roundAlt},
    \label{eq:mass-growth}
\end{align}
and under the greedy rule $\gI_\roundAlt > 0$ for every $\roundAlt$.
\end{lemma}
\begin{proof}
By definition of $\gI_\roundAlt$, we have $\predictive(\symy^{(\roundAlt)} \mid \symx^{(\roundAlt)}, \dataset_{\roundAlt-1}) \le 2^{-\gI_\roundAlt}$, so \cref{eq:bits} gives $-\log_2 \pushforward(\equivClass{\targetFunc} \mid \dataset_0) \ge \sum_{\roundAlt=1}^{\round} \gI_\roundAlt - \log_2 \pushforward(\equivClass{\targetFunc} \mid \dataset_\round)$, which rearranges to \cref{eq:mass-growth}. The greedy rule gives $\gI_\roundAlt > 0$ by \cref{lem:greedy}.
\end{proof}
\thmworst*
\begin{proof}
If $\sum_{\roundAlt \le \round} \gI_\roundAlt > \log_2 \frac{1}{2\pushforward(\equivClass{\targetFunc} \mid \dataset_0)}$, then \cref{lem:mass-growth} gives $\pushforward(\equivClass{\targetFunc} \mid \dataset_\round) > \tfrac12$ and the algorithm has terminated by round $\round$. Now let $\round$ be the right-hand side of \cref{eq:T-worst}, which is the smallest integer with $\round\gI > \log_2 \frac{1}{2\pushforward(\equivClass{\targetFunc} \mid \dataset_0)}$. If $\numRounds > \round$, then $\sum_{\roundAlt \le \round} \gI_\roundAlt \ge \round\gI$ and the algorithm would have terminated by round $\round$, a contradiction. For correctness, by \cref{eq:posterior} the posterior mass of every class consistent with $\dataset_\numRounds$ is its prior mass divided by the same normalizer, so consistent classes are ordered under the posterior as under the prior. If $\hat\classvar \ne \equivClass{\targetFunc}$, then $\pushforward(\hat\classvar \mid \dataset_\numRounds) > \tfrac12 > \pushforward(\equivClass{\targetFunc} \mid \dataset_\numRounds)$ with both classes consistent, so $\hat\classvar$ has strictly larger prior mass. Conversely, if a consistent class $\classvar'$ has larger prior mass than $\equivClass{\targetFunc}$, then $\pushforward(\classvar' \mid \dataset_\numRounds) > \pushforward(\equivClass{\targetFunc} \mid \dataset_\numRounds)$, and $\hat\classvar = \equivClass{\targetFunc}$ would give a total mass above one.
\end{proof}

\section{Experimental Details}\label{app:experimental-details}

\paragraph{Infrastructure.} All models are served using vLLM\footnote{https://docs.vllm.ai/en/}, with tensor parallelism and GPU memory scaled to model size. Models up to 8B parameters were run on a single GPU, while 32B models used either four 24 GB GPUs in parallel or a single 96 GB GPU. All servers run vLLM's batch-invariant kernels, so that seeded sampling is reproducible under concurrent load.
Models are loaded in bfloat16.
The algorithm's sampling and query-selection loop runs on a separate CPU-only node and communicates with the model server over an SSH tunnel.

\paragraph{Model parameters.} We use $\nParticles = 64$ particles, an evaluation set of size $\evalSize = 64$, resample when the effective sample size falls below $\essThreshold = 4$, and terminate once a class holds more than half of the posterior mass. Runs that return a program observe $2.3$ to $2.7$ input--output examples on average, depending on the model. The query budget we use and report is $30$ input--output pairs, which includes the seed invocations. The evaluation set is shared across models. For each problem, we draw the evaluation set once from the input distribution $\inputPrior$, instantiated with Llama3.1-8B at temperature $1.1$, conditioned on all $10$ seed invocations of the benchmark. The evaluation set therefore draws on seed examples that the algorithm does not see at $1$ seed invocation. Its inputs coincide with a seed input in $0.3\%$ of the entries, and rebuilding it from the $5$ visible seed examples changed the accuracy of Llama3.2-3B, Llama3.1-8B, and Qwen2.5-32B by at most $.02$ on $100$ problems. Rebuilding it from the single visible seed example on the full split changes accuracy by at most $.02$ on five of six models and lowers that of Qwen2.5-32B from $.342$ to $.280$ (\cref{tab:tp1}). Our method remains more accurate than the CodeARC agent on all models but Ministral-8B ($p<.01$, paired sign test) and observes fewer input--output examples on average on all but Qwen2.5-32B, as with the shared evaluation set. We keep the shared evaluation set for all settings. The guarantee of \cref{prop:pac-instance} is with respect to this input distribution. The proposal prompt and sampling temperature are selected per model on a screening set (\cref{app:prompt-selection}, \cref{fig:prompt-temperature}), and the prompts are given in \cref{app:prompts}. The maximum context length is capped at 12288 tokens across all models.

\begin{table}[!t]
  \centering
  \caption{Avg I/O and accuracy at $1$ seed invocation with the evaluation set built from all $10$ seed invocations (shared, as in \cref{tab:results}) and from the single visible one, on the full anonymized split, means over $3$ seeds. Stars test against the shared evaluation set (paired sign test, $^{**}p<.01$, $^{***}p<.001$).}
  \label{tab:tp1}
  \setlength{\aboverulesep}{0pt}\setlength{\belowrulesep}{0pt}\setlength{\tabcolsep}{4pt}
  \newcommand{\ci}[2]{#1{\scriptscriptstyle\,\pm#2}}
  \scriptsize
  \begin{tabular}{@{} l cc cc cc @{}}
    \toprule
    & \multicolumn{2}{c}{LM only (CodeARC)} & \multicolumn{2}{c}{Ours, shared} & \multicolumn{2}{c}{Ours, rebuilt from $1$ seed} \\
    \cmidrule(lr){2-3}\cmidrule(lr){4-5}\cmidrule(l){6-7}
    Model & Avg I/O & Acc. & Avg I/O & Acc. & Avg I/O & Acc. \\
    \midrule
    Llama3.2 3B & $\ci{19.18}{0.28}$ & $\ci{.129}{.023}$ & $\ci{15.93}{0.62}$ & $\ci{.169}{.022}$ & $\ci{15.96}{0.72}$ & $\ci{.157}{.016}$ \\
    Llama3.1 8B & $\ci{16.58}{0.30}$ & $\ci{.134}{.007}$ & $\ci{12.73}{1.42}$ & $\ci{.218}{.017}$ & $\ci{13.28}{0.48}^{**}$ & $\ci{.205}{.031}$ \\
    Qwen2.5 7B & $\ci{19.57}{0.31}$ & $\ci{.153}{.013}$ & $\ci{8.95}{0.94}$ & $\ci{.231}{.021}$ & $\ci{9.49}{0.67}^{**}$ & $\ci{.231}{.021}$ \\
    Olmo3 7B & $\ci{13.35}{0.11}$ & $\ci{.063}{.005}$ & $\ci{5.89}{0.64}$ & $\ci{.243}{.030}$ & $\ci{5.84}{0.30}$ & $\ci{.224}{.028}$ \\
    Ministral 8B & $\ci{16.24}{0.22}$ & $\ci{.242}{.009}$ & $\ci{13.65}{0.63}$ & $\ci{.169}{.027}$ & $\ci{14.14}{0.54}$ & $\ci{.165}{.024}$ \\
    Qwen2.5 32B & $\ci{11.13}{0.13}$ & $\ci{.248}{.025}$ & $\ci{11.14}{0.62}$ & $\ci{.342}{.027}$ & $\ci{11.30}{0.78}$ & $\ci{.280}{.012}^{***}$ \\
    \bottomrule
  \end{tabular}
\end{table}

\paragraph{Licensing.} The CodeARC benchmark \citep{wei2025codearc} is released under the Apache~2.0 license.
Llama3.1-8B and Llama3.2-3B are released under the Llama~3.1 and Llama~3.2 Community Licenses respectively. Qwen2.5-7B and Qwen2.5-32B are released under the Apache~2.0 license. Olmo3-7B is released under the Apache~2.0 license, and Ministral-8B under the Mistral Research License.

\section{Additional Results}\label{app:additional}

\subsection{Failure Modes}\label{app:failure-modes}
\begin{table}[!t]
  \centering
  \caption{Terminal state of the runs behind \cref{tab:results} at $1$ seed invocation ($\evalSize = 64$, five seeds), as a share of all runs, averaged over seeds as in \cref{tab:results}. A run converges when one class exceeds half of the posterior mass. The five rows partition the runs, and the two rows with a correct program sum to the accuracy of \cref{tab:results}.}
  \label{tab:failure-modes}
  \setlength{\tabcolsep}{3.5pt}
  \renewcommand{\arraystretch}{1.0}
  \scriptsize
  \begin{tabular}{@{} >{\raggedright\arraybackslash}p{4.3cm} cccccc @{}}
    \toprule
    & Llama3.2-3B & Llama3.1-8B & Qwen2.5-7B & Qwen2.5-32B & Olmo3-7B & Ministral-8B \\
    \midrule
    Not converged, no proposal consistent with the oracle's answers & .49 & .38 & .24 & .30 & .13 & .40 \\
    Not converged, query budget exhausted & .00 & .00 & .00 & .01 & .01 & .00 \\
    Not converged, program correct & .00 & .00 & .00 & .01 & .00 & .00 \\
    Converged, program wrong & .34 & .40 & .52 & .36 & .63 & .42 \\
    Converged, program correct & .17 & .21 & .23 & .33 & .24 & .17 \\
    \midrule
    Accuracy over all runs & .17 & .21 & .23 & .33 & .24 & .17 \\
    \bottomrule
  \end{tabular}
\end{table}
\Cref{tab:failure-modes} decomposes the runs of \cref{tab:results} by their terminal state to explain the gap between the correctness guarantee of \cref{thm:expected} and the accuracies of \cref{tab:results}. Most runs fail before the stopping rule applies. In a large share of them, between $.13$ and $.49$ depending on the model, no sampled program is consistent with the oracle's answers, so the posterior is empty. In most of the remaining runs the posterior converges on a wrong program, owing to the proposal placing its mass on a plausible alternative that survives the few queries, as expected from the miscalibrated prior of \cref{fig:prior-calibration}. Runs that keep querying until the budget is exhausted are rare, so the query budget does not limit accuracy. Both failure modes stem from the proposal. Accordingly, the particle budget (\cref{app:particles}) and the prompt (\cref{app:prompt-selection}) affect accuracy more than the query selection strategy (\cref{tab:strategies}).

\subsection{Per-Instance Bound Analysis}\label{app:bound}
\begin{table}[!t]
  \centering
  \caption{Termination round $\numRounds$ of EIG against its per-instance bound derived in \cref{thm:worst}, $\lfloor \tfrac1{\gI} \log_2 \tfrac{1}{2\pushforward(\equivClass{\targetFunc} \mid \dataset_0)} \rfloor + 1$. Runs on the diagonal, $.37$ of all runs, attain the bound.}
  \label{tab:bound}
  \setlength{\tabcolsep}{6pt}
  \renewcommand{\arraystretch}{1.0}
  \scriptsize
  \begin{tabular}{@{} l cccccc @{}}
    \toprule
    & \multicolumn{6}{c}{Bound} \\
    \cmidrule(lr){2-7}
    $\numRounds$ & $1$ & $2$ & $3$ & $4$ & $5$ & $\ge 6$ \\
    \midrule
    $1$ & 777 & 387 & 182 & 108 & 93 & 331 \\
    $2$ & -- & 106 & 112 & 60 & 46 & 94 \\
    $3$ & -- & -- & 19 & 28 & 18 & 40 \\
    $4$ & -- & -- & -- & 1 & 3 & 18 \\
    $5$ & -- & -- & -- & -- & -- & 3 \\
    \bottomrule
  \end{tabular}
\end{table}
\Cref{tab:bound} compares, run by run, the termination round of EIG with the bound of \cref{thm:worst} evaluated on that run. The bound is attained in $.37$ of the runs, and is loose when the guaranteed information is not uniform over the queries of a run, since $\gI$ is the minimum over the run's queries and a single uninformative query loosens the bound.

\subsection{Query Selection Strategies}
\cref{tab:strategies} reports, per model, the quantities of \cref{fig:turn-results} at the end of the run for every query selection strategy of \cref{alg:greedy-ig} and for the CodeARC agent, together with the average information the selected queries are guaranteed to yield. All strategies score the same candidate inputs under the same weighted particle posterior and differ only in the score. EIG maximizes the estimated information gain about the equivalence class, \cref{eq:info-gain}. Uncertainty sampling maximizes the number of distinct outputs predicted by the particles of nonzero weight. Diversity maximizes the smallest character-trigram distance, one minus the Jaccard similarity, to the inputs already queried. Random draws an input uniformly. The strategies differ in the guaranteed information. EIG guarantees close to one bit per query, uncertainty sampling somewhat less, and diversity and random selection about half as much. This ordering carries over to the number of input--output examples, where the informed strategies save about one example per run against random selection, but not to accuracy, where all four strategies differ by at most $.06$ for every model. The accuracy of a run is decided by whether the proposal contains a correct program (\cref{app:failure-modes}), which every strategy then finds within a few queries. The information criterion therefore saves oracle calls at the same accuracy. Against the CodeARC agent, it improves both for all models but Ministral-8B.\looseness=-1

\begin{table}[!t]
  \centering
  \caption{Query selection strategies of \cref{alg:greedy-ig} and the CodeARC agent with $1$ seed invocation on the full anonymized CodeARC split. Prior majority returns a program from the majority class of the initial particles without calling the oracle. Entries are means over $5$ independent seeds ($3$ for prior majority) with 95\% confidence intervals over seeds. Avg~I/O is the average number of input--output examples observed, including the seed invocation, out of a budget of $30$. Avg~$\gI_\round$ is the information the selected query is guaranteed to yield, averaged over all queries of the strategy.}
  \label{tab:strategies}
  \setlength{\aboverulesep}{0pt}
  \setlength{\belowrulesep}{0pt}
  \setlength{\tabcolsep}{4pt}
  \renewcommand{\arraystretch}{1.0}
  \newcommand{\ci}[2]{#1{\scriptscriptstyle\,\pm#2}}
  \scriptsize
  \begin{tabular}{@{} ll ccc @{}}
    \toprule
    {Model} & {Method} & {Avg I/O $\downarrow$} & {Acc. $\uparrow$} & {Avg $\gI_\round$ (bits) $\uparrow$} \\
    \midrule
    \multirow{6}{*}{Llama3.2 3B}
      & LM only (CodeARC) & $\ci{19.03}{0.44}$ & $\ci{.134}{.014}$ & -- \\
      & Prior majority (no query) & -- & $\ci{.146}{.017}$ & -- \\
      & \cellcolor{gray!10} EIG & \cellcolor{gray!10} $\ci{15.81}{0.36}$ & \cellcolor{gray!10} $\mathbf{\ci{.167}{.010}}$ & \cellcolor{gray!10} $\mathbf{\ci{.697}{.008}}$ \\
      & Uncertainty & $\mathbf{\ci{15.76}{0.18}}$ & $\ci{.165}{.007}$ & $\ci{.590}{.010}$ \\
      & Diversity & $\ci{15.96}{0.22}$ & $\ci{.166}{.008}$ & $\ci{.360}{.004}$ \\
      & Random & $\ci{16.15}{0.30}$ & $\ci{.158}{.016}$ & $\ci{.362}{.022}$ \\
    \midrule
    \multirow{6}{*}{Llama3.1 8B}
      & LM only (CodeARC) & $\ci{16.59}{0.13}$ & $\ci{.130}{.015}$ & -- \\
      & Prior majority (no query) & -- & $\ci{.179}{.015}$ & -- \\
      & \cellcolor{gray!10} EIG & \cellcolor{gray!10} $\mathbf{\ci{12.88}{0.55}}$ & \cellcolor{gray!10} $\mathbf{\ci{.212}{.011}}$ & \cellcolor{gray!10} $\mathbf{\ci{.765}{.039}}$ \\
      & Uncertainty & $\ci{13.05}{0.51}$ & $\ci{.210}{.010}$ & $\ci{.640}{.028}$ \\
      & Diversity & $\ci{13.33}{0.43}$ & $\ci{.204}{.013}$ & $\ci{.387}{.016}$ \\
      & Random & $\ci{13.78}{0.37}$ & $\ci{.200}{.011}$ & $\ci{.382}{.014}$ \\
    \midrule
    \multirow{6}{*}{Qwen2.5 7B}
      & LM only (CodeARC) & $\ci{19.55}{0.25}$ & $\ci{.149}{.009}$ & -- \\
      & Prior majority (no query) & -- & $\ci{.168}{.006}$ & -- \\
      & \cellcolor{gray!10} EIG & \cellcolor{gray!10} $\mathbf{\ci{9.00}{0.42}}$ & \cellcolor{gray!10} $\mathbf{\ci{.230}{.008}}$ & \cellcolor{gray!10} $\mathbf{\ci{.796}{.019}}$ \\
      & Uncertainty & $\ci{9.24}{0.38}$ & $\ci{.229}{.008}$ & $\ci{.676}{.022}$ \\
      & Diversity & $\ci{9.52}{0.20}$ & $\ci{.213}{.010}$ & $\ci{.433}{.004}$ \\
      & Random & $\ci{9.95}{0.27}$ & $\ci{.213}{.008}$ & $\ci{.448}{.009}$ \\
    \midrule
    \multirow{6}{*}{Qwen2.5 32B}
      & LM only (CodeARC) & $\ci{11.12}{0.06}$ & $\ci{.250}{.011}$ & -- \\
      & Prior majority (no query) & -- & $\ci{.217}{.016}$ & -- \\
      & \cellcolor{gray!10} EIG & \cellcolor{gray!10} $\mathbf{\ci{10.84}{0.32}}$ & \cellcolor{gray!10} $\ci{.335}{.023}$ & \cellcolor{gray!10} $\mathbf{\ci{.722}{.079}}$ \\
      & Uncertainty & $\ci{11.08}{0.28}$ & $\mathbf{\ci{.378}{.035}}$ & $\ci{.608}{.050}$ \\
      & Diversity & $\ci{11.07}{0.36}$ & $\ci{.316}{.005}$ & $\ci{.313}{.025}$ \\
      & Random & $\ci{11.59}{0.59}$ & $\ci{.353}{.020}$ & $\ci{.337}{.046}$ \\
    \midrule
    \multirow{6}{*}{Olmo3 7B}
      & LM only (CodeARC) & $\ci{13.35}{0.16}$ & $\ci{.063}{.007}$ & -- \\
      & Prior majority (no query) & -- & $\ci{.159}{.031}$ & -- \\
      & \cellcolor{gray!10} EIG & \cellcolor{gray!10} $\mathbf{\ci{5.90}{0.33}}$ & \cellcolor{gray!10} $\ci{.241}{.014}$ & \cellcolor{gray!10} $\mathbf{\ci{.990}{.026}}$ \\
      & Uncertainty & $\ci{6.01}{0.31}$ & $\mathbf{\ci{.249}{.027}}$ & $\ci{.856}{.036}$ \\
      & Diversity & $\ci{6.58}{0.34}$ & $\ci{.235}{.016}$ & $\ci{.521}{.005}$ \\
      & Random & $\ci{7.13}{0.29}$ & $\ci{.243}{.037}$ & $\ci{.519}{.019}$ \\
    \midrule
    \multirow{6}{*}{Ministral 8B}
      & LM only (CodeARC) & $\ci{16.22}{0.12}$ & $\mathbf{\ci{.239}{.013}}$ & -- \\
      & Prior majority (no query) & -- & $\ci{.136}{.020}$ & -- \\
      & \cellcolor{gray!10} EIG & \cellcolor{gray!10} $\mathbf{\ci{13.53}{0.27}}$ & \cellcolor{gray!10} $\ci{.171}{.010}$ & \cellcolor{gray!10} $\mathbf{\ci{.791}{.026}}$ \\
      & Uncertainty & $\ci{13.63}{0.17}$ & $\ci{.170}{.003}$ & $\ci{.679}{.023}$ \\
      & Diversity & $\ci{14.00}{0.34}$ & $\ci{.162}{.011}$ & $\ci{.389}{.031}$ \\
      & Random & $\ci{14.67}{0.50}$ & $\ci{.156}{.006}$ & $\ci{.385}{.027}$ \\
    \bottomrule
  \end{tabular}
\end{table}

\subsection[Controls and Literature Baselines]{Controls and Literature Baselines}\label{app:controls}
\Cref{tab:weights,tab:ports,tab:pv} separate the contributions of the importance weights, the selection rule, and the sampling budget. All runs use $1$ seed invocation and $3$ seeds, and stars mark a paired sign test over problems and seeds ($^{**}p<.01$, $^{***}p<.001$). \Cref{tab:weights} replaces the importance weights by uniform weights over consistent programs, as in \citet{kobalczyk2025active}, and by the exact weights of the sampled completions. The latter requires sampling at temperature $1$, so all three arms use it here. The approximate weights of \cref{sec:snis} and the exact weights perform on par, and both require significantly fewer input--output examples than uniform weights for all six models, at no significant difference in accuracy. \Cref{tab:ports} ports the selection rules of \citet{ji2020question} and \citet{lee2025program} (minimax with uniform weights) and of \citet{kobalczyk2025active} (EIG with uniform weights) into \cref{alg:greedy-ig}, keeping the proposal, evaluation set, and stopping rule fixed. Both require significantly more input--output examples than our method, at no significant difference in accuracy. \Cref{tab:pv} matches the sampling budget. A propose-and-verify baseline draws $112$ programs, the average number our method samples, keeps those consistent with the seed example, and returns a program of the majority class. Our method is significantly more accurate for five of six models.

\begin{table}[!t]
  \centering
  \caption{Importance weights against uniform weights, on $100$ problems at sampling temperature $1$. Stars test against uniform weights.}
  \label{tab:weights}
  \setlength{\aboverulesep}{0pt}\setlength{\belowrulesep}{0pt}\setlength{\tabcolsep}{4pt}
  \newcommand{\ci}[2]{#1{\scriptscriptstyle\,\pm#2}}
  \scriptsize
  \begin{tabular}{@{} l ccc ccc @{}}
    \toprule
    & \multicolumn{3}{c}{Avg I/O $\downarrow$} & \multicolumn{3}{c}{Acc. $\uparrow$} \\
    \cmidrule(lr){2-4}\cmidrule(lr){5-7}
    Model & Uniform & Approximate & Exact & Uniform & Approximate & Exact \\
    \midrule
    Llama3.2 3B & $\ci{18.24}{2.12}$ & $\ci{16.68}{2.63}^{***}$ & $\ci{16.67}{2.64}^{***}$ & $\ci{.133}{.051}$ & $\ci{.143}{.025}$ & $\ci{.139}{.029}$ \\
    Llama3.1 8B & $\ci{14.87}{0.73}$ & $\ci{13.10}{0.54}^{***}$ & $\ci{12.89}{0.84}^{***}$ & $\ci{.167}{.064}$ & $\ci{.173}{.000}$ & $\ci{.180}{.015}$ \\
    Qwen2.5 7B & $\ci{11.04}{1.75}$ & $\ci{10.07}{2.20}^{***}$ & $\ci{10.08}{2.19}^{***}$ & $\ci{.187}{.029}$ & $\ci{.163}{.101}$ & $\ci{.180}{.073}$ \\
    Olmo3 7B & $\ci{8.75}{1.62}$ & $\ci{6.28}{1.10}^{***}$ & $\ci{6.27}{1.12}^{***}$ & $\ci{.256}{.046}$ & $\ci{.229}{.038}$ & $\ci{.239}{.061}$ \\
    Ministral 8B & $\ci{14.37}{1.00}$ & $\ci{13.21}{1.21}^{***}$ & $\ci{13.07}{1.53}^{***}$ & $\ci{.197}{.082}$ & $\ci{.193}{.090}$ & $\ci{.159}{.066}$ \\
    Qwen2.5 32B & $\ci{10.54}{1.22}$ & $\ci{9.18}{1.57}^{***}$ & $\ci{9.18}{1.57}^{***}$ & $\ci{.296}{.141}$ & $\ci{.303}{.106}$ & $\ci{.282}{.102}$ \\
    \bottomrule
  \end{tabular}
\end{table}

\begin{table}[!t]
  \centering
  \caption{Selection rules from the literature ported into \cref{alg:greedy-ig}, on $100$ problems at the sampling temperatures of the main experiments. Stars test against our method.}
  \label{tab:ports}
  \setlength{\aboverulesep}{0pt}\setlength{\belowrulesep}{0pt}\setlength{\tabcolsep}{4pt}
  \newcommand{\ci}[2]{#1{\scriptscriptstyle\,\pm#2}}
  \scriptsize
  \begin{tabular}{@{} l ccc ccc @{}}
    \toprule
    & \multicolumn{3}{c}{Avg I/O $\downarrow$} & \multicolumn{3}{c}{Acc. $\uparrow$} \\
    \cmidrule(lr){2-4}\cmidrule(lr){5-7}
    Model & Ours & Minimax, unif. & EIG, unif. & Ours & Minimax, unif. & EIG, unif. \\
    \midrule
    Llama3.2 3B & $\ci{15.49}{3.27}$ & $\ci{16.37}{3.33}^{**}$ & $\ci{15.84}{2.65}^{**}$ & $\ci{.153}{.076}$ & $\ci{.146}{.077}$ & $\ci{.170}{.053}$ \\
    Llama3.1 8B & $\ci{13.49}{1.92}$ & $\ci{14.01}{2.90}^{**}$ & $\ci{14.09}{0.96}^{***}$ & $\ci{.185}{.058}$ & $\ci{.195}{.063}$ & $\ci{.182}{.090}$ \\
    Qwen2.5 32B & $\ci{9.55}{3.13}$ & $\ci{10.45}{3.71}^{***}$ & $\ci{11.60}{1.31}^{***}$ & $\ci{.263}{.017}$ & $\ci{.293}{.086}$ & $\ci{.290}{.103}$ \\
    \bottomrule
  \end{tabular}
\end{table}

\begin{table}[!t]
  \centering
  \caption{Accuracy of a compute-matched propose-and-verify baseline ($112$ sampled programs, no oracle queries) and of our method on the full anonymized split. Stars test against propose-and-verify.}
  \label{tab:pv}
  \setlength{\aboverulesep}{0pt}\setlength{\belowrulesep}{0pt}\setlength{\tabcolsep}{4pt}
  \newcommand{\ci}[2]{#1{\scriptscriptstyle\,\pm#2}}
  \scriptsize
  \begin{tabular}{@{} l cc @{}}
    \toprule
    Model & Propose-and-verify & Ours \\
    \midrule
    Llama3.2 3B & $\ci{.159}{.019}$ & $\ci{.169}{.022}$ \\
    Llama3.1 8B & $\ci{.183}{.018}$ & $\ci{.219}{.015}^{***}$ \\
    Qwen2.5 7B & $\ci{.175}{.010}$ & $\ci{.232}{.018}^{***}$ \\
    Olmo3 7B & $\ci{.160}{.032}$ & $\ci{.243}{.031}^{***}$ \\
    Ministral 8B & $\ci{.148}{.010}$ & $\ci{.169}{.028}^{**}$ \\
    Qwen2.5 32B & $\ci{.224}{.011}$ & $\ci{.342}{.028}^{***}$ \\
    \bottomrule
  \end{tabular}
\end{table}

\subsection{Particle Budget}\label{app:particles}
\begin{figure}[!t]
    \centering
    \includegraphics[width=\linewidth]{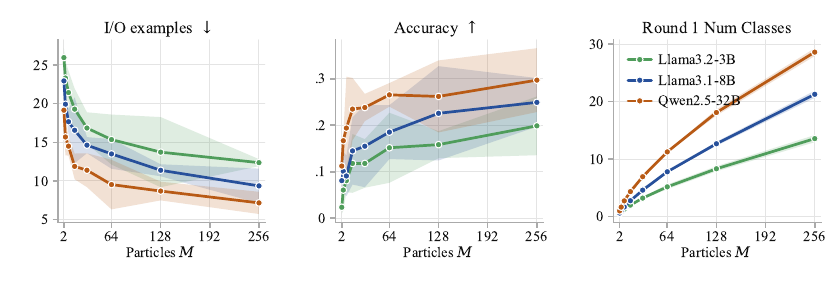}
    \caption{Particle-budget ablation with $1$ seed invocation on $100$ problems. We show the input--output examples observed, the accuracy, and the number of behavioral classes at round $1$ against the number of particles $\nParticles$, with 95\% confidence intervals over $3$ seeds.}
    \label{fig:particles}
\end{figure}

\cref{fig:particles} varies the number of particles $\nParticles$ from $2$ to $256$ for three models on $100$ problems with $1$ seed invocation. More particles produce more behavioral classes at round $1$, so the first query is more informative, and the algorithm ends with fewer input--output examples and higher accuracy. Accuracy keeps rising past the $\nParticles = 64$ of the main experiments and does not saturate at $256$, which agrees with the failure modes of \cref{app:failure-modes}, since the runs that converge on a wrong program or on none are runs in which no correct program was sampled, and a larger particle set samples one more often. We use $64$ to limit the per-round cost, which grows with the number of programs that have to be executed on $\testSet$ (\cref{tab:cost}).

\subsection{Proposal Prompt and Sampling Temperature}\label{app:prompt-selection}
\begin{figure}[!t]
    \centering
    \includegraphics[width=\linewidth]{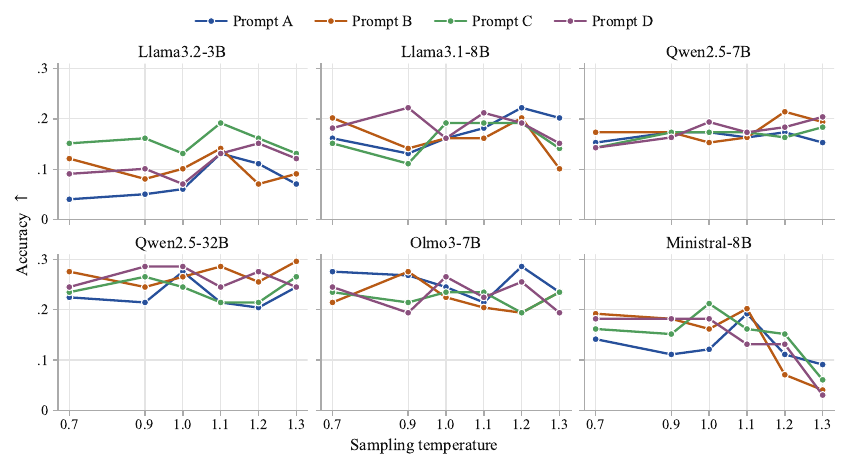}
    \caption{Accuracy of our algorithm against the sampling temperature for each proposal prompt, with $1$ seed invocation on $100$ problems. Prompts are ordered by accuracy pooled over \mbox{models and temperatures}.}
    \label{fig:prompt-temperature}
\end{figure}

The proposal $\proposalProg_\round$ depends on the prompt shown to the LM and on its sampling temperature. We consider four prompts (Prompts A--D, \cref{app:prompts}), which differ in how strongly they ask for hypotheses that disagree on unseen inputs, and six temperatures between $0.7$ and $1.3$. We choose the prompt and temperature for each model in a preliminary screen over prompts and temperatures in this range. The $100$ screening problems are part of the evaluation split. Recomputing \cref{tab:results} on the remaining $985$ problems changes every entry by at most $.007$. \Cref{fig:prompt-temperature} reports the accuracy of the full algorithm over the grid of prompts and temperatures. No prompt is best for every model. With about $100$ problems per configuration, differences of a few hundredths are within sampling noise. Ministral-8B degrades at temperatures above $1.1$ and Llama3.2-3B performs worst with the shortest prompt (Prompt A), with accuracy varying by $.18$ and $.15$ across the grid for these two models, compared with $.07$ to $.12$ for the others.

\section{Prompts}\label{app:prompts}
We show each prompt as the text the model receives, with placeholders for problem-specific content written as \textcolor{promptph}{\texttt{<NAME>}}. All prompts are sent as a user message.

\subsection{Default CodeARC Prompts}
The baseline is a multi-turn conversation. Prompt B1 opens it. Depending on the model's previous action, Prompt B2 or B3 is appended as the next user message, and Prompt B4 closes the conversation.

\begin{promptbox}{Prompt B1: first turn}
You are a software engineer tasked with implementing a blackbox Python function `<FUNCTION_NAME>`. Your goal is to implement this function correctly, ensuring it works for ALL possible valid inputs, not just the example inputs provided.

Your primary goal is to get the function implementation exactly right.
Your secondary goal is to minimize the number of function invocations and implementation attempts.

GIVEN EXAMPLES:
Function invocations:
<INPUT_INVOCATIONS>

Corresponding outputs:
<OUTPUT_INVOCATIONS>

STRATEGY OVERVIEW:
You should work in two phases:
1. EXPLORATION PHASE: Use function invocations strategically to fully understand the function's behavior
2. IMPLEMENTATION PHASE: Only after thorough understanding, attempt an implementation

This is the first iteration. You must decide between one of the following two options:
  1. Generate more function invocations in the same format as the given examples. Choose this option if you are not confident about how the function works. Each invocation should be strategically designed to:
     - Test a specific hypothesis about the function's behavior
     - Explore edge cases (empty inputs, negative numbers, special characters, etc.)
     - Refine your understanding of patterns in the output
     - Verify your assumptions systematically

    You can generate up to <MAX_INVOCATIONS> function invocations across all iterations. You must generate exactly 1 invocation in this iteration.

    You should first reason about the problem and the invocations to generate. Then, return the function invocations in the following format:
    INVOCATIONS:
    ```python
    print('Result i: ' + str(<FUNCTION_NAME>(args)))
    print('Result i+1: ' + str(<FUNCTION_NAME>(args)))
    ...
    ```
    where each print statement is on a new line. ONLY include executable Python code within the code block. You must specify "INVOCATIONS" in the response.

  2. Generate an implementation of `<FUNCTION_NAME>`.

    You have only <MAX_CHECKS> debugging checks total. If your implementation is incorrect, you lose one check and are shown a counterexample. If you exhaust all checks with incorrect implementations, you CANNOT take any more actions and your final attempt will be considered your answer.

    You should ONLY choose this option if ALL of the following conditions are met:
    - You have seen enough diverse examples to understand the function completely
    - You have verified your understanding through strategic invocations
    - You have tested edge cases and corner cases thoroughly
    - You can confidently define the function for ALL possible valid inputs
    - You have a high degree of certainty about your implementation

    If you choose this option, you should first reason about the problem and the implementation to generate. Then, return the implementation in the following format:
    IMPLEMENTATION:
    ```python
    def <FUNCTION_NAME>(input1, input2, ...):
      ...
    ```
    ONLY include executable Python code within the code block. You must specify "IMPLEMENTATION" in the response.

Reason about which of the two actions described above to perform and then ONLY perform one of the two actions, specifying which action ("INVOCATIONS" or "IMPLEMENTATION") you performed in the response.
\end{promptbox}

\begin{promptbox}{Prompt B2: next turn, after the model requested invocations}
Here are the outputs of the function invocations from the previous iteration:
<OUTPUT_INVOCATIONS>

STRATEGY UPDATE:
Based on the new data above, you should update your understanding of the function and decide on your next action.

Following the formats described previously, you must decide between one of the two following options:
  1. Generate more function invocations to further explore the function's behavior. You have <MAX_INVOCATIONS> function invocations left across all remaining iterations. You must generate exactly 1 invocation in this iteration.
     [...same INVOCATIONS format block as Prompt B1...]

  2. Generate an implementation of `<FUNCTION_NAME>`.
     You have <MAX_CHECKS> debugging checks left.
     [...same IMPLEMENTATION format block as Prompt B1...]

Reason about which of the two actions described above to perform and then ONLY perform one of the two actions, specifying which action ("INVOCATIONS" or "IMPLEMENTATION") you performed in the response.
\end{promptbox}

\begin{promptbox}{Prompt B3: next turn, after a submitted implementation failed}
The implementation you generated in the previous iteration failed for the following input (and potentially other inputs):
<FAILED_INPUTS>

Your implementation was incorrect. You now have <MAX_CHECKS> debugging checks remaining.

This failure provides valuable information about a case you didn't account for. You must decide between:

  1. Generate more function invocations to better understand why your implementation failed. You have <MAX_INVOCATIONS> function invocations left across all remaining iterations. You must generate exactly 1 invocation in this iteration.
     After a failed implementation, generating targeted invocations is almost always the best strategy; focus on:
     - Inputs similar to the failed case to understand the pattern you missed
     - Variations of the failed input to isolate the exact behavior
     - Edge cases related to the failure point
     - Inputs that can help discriminate between competing hypotheses
     [...same INVOCATIONS format block as Prompt B1...]

  2. Generate a revised implementation of `<FUNCTION_NAME>`.
     You have <MAX_CHECKS> debugging checks. Only choose this option if you are absolutely certain you understand why your previous implementation failed AND you are confident your new implementation addresses that issue AND accounts for all other possible inputs.
     [...same IMPLEMENTATION format block as Prompt B1...]

Reason about which of the two actions described above to perform and then ONLY perform one of the two actions, specifying which action ("INVOCATIONS" or "IMPLEMENTATION") you performed in the response.
\end{promptbox}

\begin{promptbox}{Prompt B4: final turn, once all debugging checks are used}
Here are the outputs of the function invocations from the previous iteration:
<OUTPUT_INVOCATIONS>

Now, synthesize all the information you've gathered across all iterations. Analyze patterns, edge cases, and the function's overall behavior.

Generate your final implementation of the function. Make sure it handles all cases, including the ones you've explicitly tested and others that follow the same pattern.

Return the final implementation in the following format:
```python
def <FUNCTION_NAME>(input1, input2, ...):
  ...
```
ONLY include executable Python code within the code block.
\end{promptbox}

\subsection{Proposal Prompts}
These are the four proposal prompts of our algorithm, and \cref{app:prompt-selection} describes which one is used for each model. The examples displayed are the seed invocations and all queries answered so far. The program prior uses the same prompt, displaying only the seed invocations.

\begin{promptbox}{Prompt A}
Implement the hidden Python function `<FUNCTION_NAME>`.

Function invocations:
<INPUT_INVOCATIONS>

Corresponding outputs:
<OUTPUT_INVOCATIONS>

Return exactly one implementation using only the structure below without any additional explanations:
IMPLEMENTATION:
```python
def <FUNCTION_NAME>(input1, input2, ...):
  ...
```
\end{promptbox}

\begin{promptbox}{Prompt B}
Implement the hidden Python function `<FUNCTION_NAME>` from the examples below.

GIVEN EXAMPLES:
Function invocations:
<INPUT_INVOCATIONS>

Corresponding outputs:
<OUTPUT_INVOCATIONS>

First, list concretely HOW the examples underdetermine the function: name specific inputs NOT shown above on which two valid implementations would disagree (edge cases, empty inputs, ties, ordering, types, unusual characters). Then pick ONE unresolved interpretation -- not necessarily the most popular one -- and implement it. The implementation must reproduce all given examples exactly. Return exactly one implementation using only the structure below without any additional explanations:
IMPLEMENTATION:
```python
def <FUNCTION_NAME>(input1, input2, ...):
  ...
```
\end{promptbox}

\begin{promptbox}{Prompt C}
You are a software engineer tasked with implementing a blackbox Python function `<FUNCTION_NAME>` from a set of input-output pairs provided below.
Your goal is to implement this function correctly, ensuring it works for ALL possible valid inputs, not just the example inputs provided.

GIVEN EXAMPLES:
Function invocations:
<INPUT_INVOCATIONS>

Corresponding outputs:
<OUTPUT_INVOCATIONS>

You should first briefly reason about the problem and all possible implementations that may lead to this input-output behavior. Every proposed implementation must satisfy all provided input-output examples exactly. Across proposals, favor diversity in hypotheses and complexity: include simpler and more structured alternatives (for example direct formulas, conditionals, loops, helper functions, or different algorithmic strategies), while staying consistent with the examples. Return exactly one implementation using only the structure below without any additional explanations:
IMPLEMENTATION:
```python
def <FUNCTION_NAME>(input1, input2, ...):
  ...
```
\end{promptbox}

\begin{promptbox}{Prompt D}
You are a software engineer tasked with implementing a blackbox Python function `<FUNCTION_NAME>` from a set of input-output pairs provided below.
Your goal is to implement this function correctly, ensuring it works for ALL possible valid inputs, not just the example inputs provided.

GIVEN EXAMPLES:
Function invocations:
<INPUT_INVOCATIONS>

Corresponding outputs:
<OUTPUT_INVOCATIONS>

IMPORTANT: this same prompt is being sent to you many times independently, and your answers are pooled into a set of competing hypotheses. Do NOT default to the single most obvious implementation every time. Treat your answer as ONE DRAW from the space of all distinct rules that reproduce the given examples exactly, sampled in proportion to how plausible each rule is. Different draws should disagree with each other on inputs that are NOT in the given examples, while all agreeing on the examples themselves.

You should first briefly reason about the problem and enumerate several genuinely different rules that all reproduce the given input-output behavior -- they may differ in how they generalise (edge cases, empty or negative or extreme inputs, types, ordering, off-by-one boundaries) or in algorithmic strategy (direct formula, conditionals, loops, recursion, helper functions, library calls). Then pick ONE of them and commit to it. Every proposed implementation must satisfy all provided input-output examples exactly. Return exactly one implementation using only the structure below without any additional explanations:
IMPLEMENTATION:
```python
def <FUNCTION_NAME>(input1, input2, ...):
  ...
```
\end{promptbox}

\subsection{Input Distribution Prompt}
This prompt instantiates the input distribution $\inputPrior$. It is sampled repeatedly, once per problem and before the first query, and each parsed completion becomes one element of $\testSet$.

\begin{promptbox}{Input distribution prompt}
You are choosing one informative test input for a hidden Python function `<FUNCTION_NAME>`.

Known examples:
Function invocations:
<INPUT_INVOCATIONS>

Corresponding outputs:
<OUTPUT_INVOCATIONS>

Reply with ONLY the INVOCATIONS block below--no preamble, no headings, no explanations, and no text outside the fenced Python block.

INVOCATIONS:
```python
print('Result 1: ' + str(solution(args)))
```

Prefer diverse inputs (edges, negatives, extremes, unusual structures) inside the single print(...) line.
\end{promptbox}

\section{AI Use Statement}\label{app:ai-use}
We used generative AI tools for brainstorming, to polish writing and mathematical proofs, and to find some of the related work. We reviewed all AI-generated text and retrieved resources and take full responsibility for the content of the paper.

\end{document}